\documentclass{article}
\usepackage[margin=1.2in,letterpaper]{geometry}
\usepackage{hyperref}
\usepackage{url}
\usepackage{multirow}
\usepackage{authblk}
\usepackage{pdflscape}

\usepackage{amsmath,amsbsy,amsfonts,amssymb,amsthm,dsfont}
\usepackage{cite,xspace}
\usepackage{algorithm,algorithmic}
\usepackage{cleveref}
\usepackage{graphicx}
\usepackage{rotating}

\def\ddefloop#1{\ifx\ddefloop#1\else\ddef{#1}\expandafter\ddefloop\fi}
\def\ddef#1{\expandafter\def\csname b#1\endcsname{\ensuremath{\mathbf{#1}}}}
\ddefloop ABCDEFGHIJKLMNOPQRSTUVWXYZ\ddefloop
\def\ddef#1{\expandafter\def\csname bb#1\endcsname{\ensuremath{\mathbb{#1}}}}
\ddefloop ABCDEFGHIJKLMNOPQRSTUVWXYZ\ddefloop
\def\ddef#1{\expandafter\def\csname c#1\endcsname{\ensuremath{\mathcal{#1}}}}
\ddefloop ABCDEFGHIJKLMNOPQRSTUVWXYZ\ddefloop
\def\ddef#1{\expandafter\def\csname v#1\endcsname{\ensuremath{\boldsymbol{#1}}}}
\ddefloop ABCDEFGHIJKLMNOPQRSTUVWXYZabcdefghijklmnopqrstuvwxyz\ddefloop
\def\ddef#1{\expandafter\def\csname
  v#1\endcsname{\ensuremath{\boldsymbol{\csname #1\endcsname}}}}
\ddefloop
    {alpha}{beta}{gamma}{delta}{epsilon}{varepsilon}{zeta}{eta}{theta}{vartheta}{iota}{kappa}{lambda}{mu}{nu}{xi}{pi}{varpi}{rho}{varrho}{sigma}{varsigma}{tau}{upsilon}{phi}{varphi}{chi}{psi}{omega}{Gamma}{Delta}{Theta}{Lambda}{Xi}{Pi}{Sigma}{varSigma}{Upsilon}{Phi}{Psi}{Omega}\ddefloop

\renewcommand\v{\ensuremath{\boldsymbol}}

\newcommand\norm[1]{\left\| #1 \right\|}

\newcommand{\parent}{\ensuremath{P}}
\newcommand{\spindex}{\ensuremath{\gamma}}

\newcommand\alg{\ensuremath{\mathtt{apple}}\xspace}

\newcommand\ip[2]{\left\langle #1,#2\right\rangle}
\def\nf{\nabla f}

\newtheorem{theorem}{Theorem}

\newcommand{\linear}{\ensuremath{\mathtt{\ell}}}
\newcommand{\quadratic}{\ensuremath{\mathtt{q}}}
\newcommand{\cubic}{\ensuremath{\mathtt{c}}}
\newcommand{\lalg}{\ensuremath{\mathtt{lin}}\xspace}
\newcommand{\qalg}{\ensuremath{\mathtt{quad}}\xspace}
\newcommand{\calg}{\ensuremath{\mathtt{cubic}}\xspace}
\newcommand{\balg}{\ensuremath{\mathtt{bigram}}\xspace}
\newcommand{\ssm}{\ensuremath{\mathtt{ssm}}\xspace}

\newcommand{\algplain}{\ensuremath{\mathtt{alg}}}

\newcommand{\rerr}{\ensuremath{\mathrm{rel\,err}}}

\newcommand{\rtime}{\ensuremath{\mathrm{rel\,time}}}
\newcommand{\runtime}{\ensuremath{\mathrm{time}}}

\newcommand{\FS}[1]{\ensuremath{S_{(#1)}}}

\title{Scalable Nonlinear Learning with \\ Adaptive Polynomial Expansions}

\author[1]{\mbox{Alekh Agarwal}}
\author[2]{\mbox{Alina Beygelzimer}}
\author[3]{\mbox{Daniel Hsu}}
\author[1]{\mbox{John Langford}}
\author[4]{\mbox{Matus Telgarsky\thanks{This work was performed while MT was visiting Microsoft Research, NYC.}}}
\affil[1]{Microsoft Research}
\affil[2]{Yahoo!\ Labs}
\affil[3]{Columbia University}
\affil[4]{Rutgers University}

\begin{document}

\maketitle

\begin{abstract}%
  Can we effectively learn a nonlinear representation in time comparable
to linear learning?  We describe a new algorithm that explicitly and
adaptively expands higher-order interaction features over base linear
representations.  The algorithm is designed for extreme computational
efficiency, and an extensive experimental study shows that its
computation/prediction tradeoff ability compares very favorably against
strong baselines.

\end{abstract}

\section{Introduction}
\label{sec:intro}
When faced with large datasets, it is commonly observed that using all
the data with a simpler algorithm is superior to using a small
fraction of the data with a more computationally intense but possibly
more effective algorithm.  The question becomes: What is the most
sophisticated algorithm that can be executed given a computational 
constraint?

At the largest scales, Na\"ive Bayes approaches offer a simple, easily
distributed single-pass algorithm.  A more computationally difficult,
but commonly better-performing approach is large scale linear 
regression, which has been effectively parallelized in several ways on
real-world large scale datasets~\cite{MCFS13,ACDL14}. Is there a
modestly more computationally difficult approach that allows us to
commonly achieve superior statistical performance?

The approach developed here starts with a fast parallelized online
learning algorithm for linear models, and explicitly
and adaptively adds higher-order interaction features over the
course of training, using the learned weights as a guide.
The resulting space of polynomial functions
increases the approximation power over the base linear
representation at a modest increase in computational cost. 

Several natural folklore baselines exist.  For example, it is common
to enrich feature spaces with $n$-grams or low-order interactions.
These approaches are naturally computationally appealing, because
these nonlinear features can be computed on-the-fly avoiding I/O
bottlenecks.  With I/O bottlenecked datasets, this can sometimes even
be done so efficiently that the additional computational complexity is
negligible, so improving over this baseline is quite challenging.

\begin{figure}
  \begin{center}
    \includegraphics[width=0.75\textwidth]{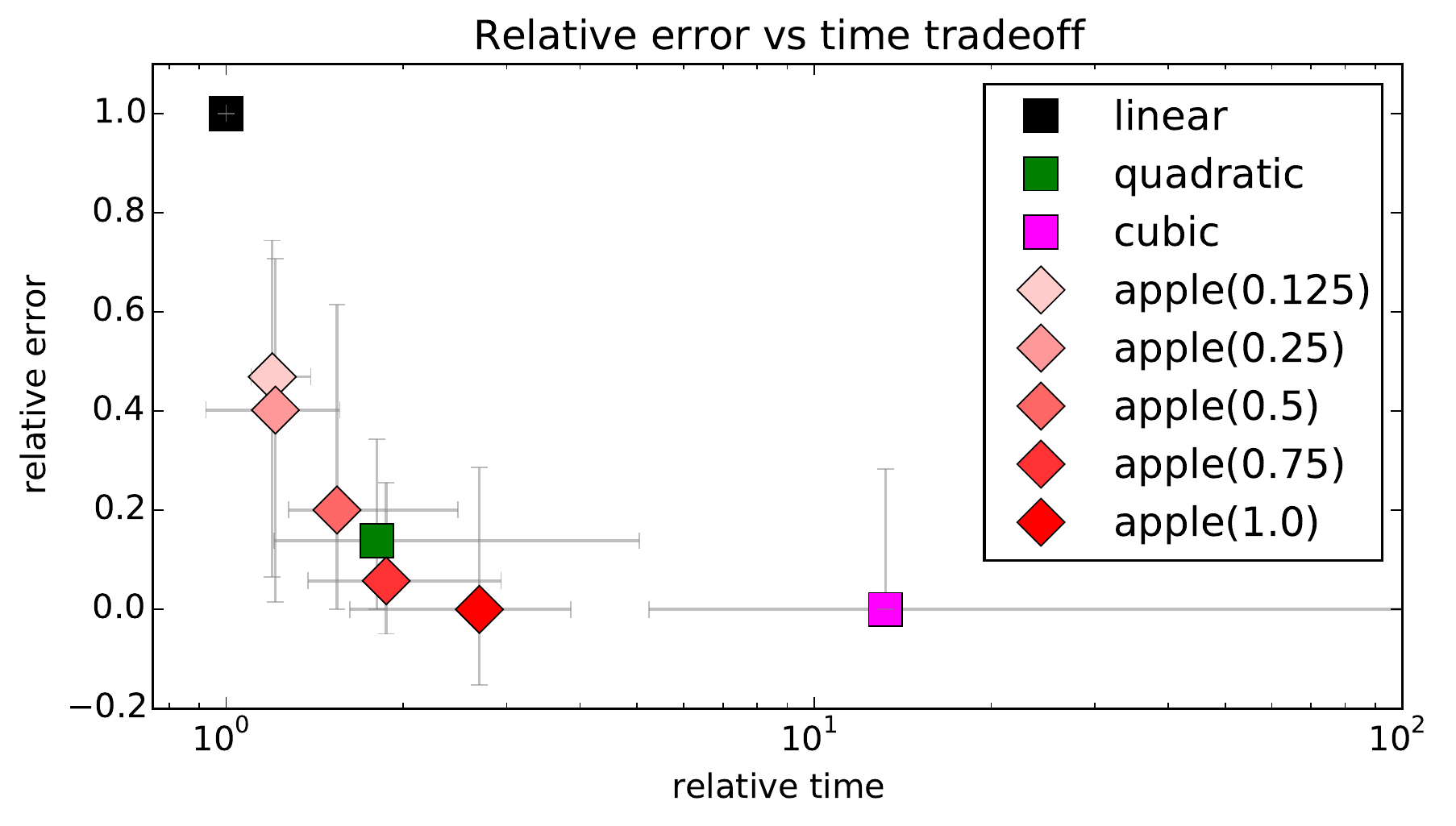}
  \end{center}
  \vspace{-6mm}
  \caption{%
    Computation/prediction tradeoff points using non-adaptive
    polynomial expansions and adaptive polynomial expansions (\alg).
    The markers are positioned at the coordinate-wise median of
    $(\mathtt{relative\ error},\mathtt{relative\ time})$ over $30$
    datasets, with bars extending to $25$th and $75$th percentiles.
    See Section~\ref{sec:experiments} for definition of relative error
    and relative time used here.}
  \label{fig:error_vs_time}
\end{figure}

The design of our algorithm
is heavily influenced by considerations for computational efficiency,
as discussed further in Section~\ref{sec:algorithm}.  Several alternative
designs are plausible but fail to provide
adequate computation/prediction tradeoffs or even outperform the
aforementioned folklore baselines.  An extensive experimental study
in Section~\ref{sec:experiments} compares efficient implementations of
these baselines with the proposed mechanism and gives strong evidence
of the latter's dominant computation/prediction tradeoff ability (see
Figure~\ref{fig:error_vs_time} for an illustrative summary).

Although it is notoriously difficult to analyze nonlinear algorithms,
it turns out that two aspects of this algorithm are amenable to
analysis.  First, we prove a regret bound showing that we can
effectively compete with a growing feature set.  Second, we exhibit
simple problems where this algorithm is effective, and discuss a
worst-case consistent variant.

\paragraph{Related work.}

This work considers methods for enabling nonlinear learning directly
in a highly-scalable learning algorithm.  Starting with a fast
algorithm is desirable because it more naturally allows one to improve
statistical power by spending more computational resources until a
computational budget is exhausted.  In contrast, many existing
techniques start with a (comparably) slow method (e.g., kernel
SVM~\cite{WS01}, batch PCA~\cite{Mahoney11}, batch
least-squares regression~\cite{Mahoney11}), and speed it up by
sacrificing statistical power, often just to allow the algorithm to
run at all on massive data sets.
Similar challenges also arise in
exploring the tradeoffs with boosting~\cite{FS97}, where typical
weak learners involve either exhaustive search or batch algorithms
(e.g., decision tree induction~\cite{Friedman99,JZ14}) that present
their own challenges in scaling and parallelization.


A standard alternative to explicit polynomial expansions is to employ
polynomial kernels with the kernel trick~\cite{SS02}.  While kernel
methods generally have computation scaling at least quadratically with
the number of training examples, a number of approximations schemes
have been developed to enable a better tradeoff.  The Nystr\"om method
(and related techniques) can be used to approximate the kernel matrix
while permitting faster training~\cite{WS01}. However, these methods
still suffer from the drawback that the model size after $n$ examples
is typically $O(n)$. As a result, even single pass online
implementations~\cite{BordesEWB2005} typically suffer from $O(n^2)$
training and $O(n)$ testing time complexity.

Another class of approximation schemes for kernel methods involves
random embeddings into a high (but finite) dimensional Euclidean space
such that the standard inner product there approximates the kernel
function~\cite{RR08,KK12,PP13,HGXD14}. Recently, such schemes have
been developed for polynomial kernels~\cite{KK12,PP13,HGXD14} with
computational scaling roughly linear in the polynomial
degree. However, for many sparse, high-dimensional datasets (such as
text data), the embedding of~\cite{PP13} creates dense, high
dimensional examples, which leads to a substantial increase in
computational complexity. Moreover, neither of the embeddings
from~\cite{KK12,PP13} exhibits good statistical performance unless
combined with dense linear dimension reduction~\cite{HGXD14}, which
again results in dense vector computations.  Such feature construction
schemes are also typically unsupervised, while the method proposed
here makes use of label information.

Learning sparse polynomial functions is primarily a computational
challenge.
This is because the na\"ive approach of combining explicit,
non-adaptive polynomial expansions with sparse regression is
statistically sound; the problem is its running time, which scales
with $d^\ell$ for degree-$\ell$ polynomials in $d$ dimensions.
Among methods proposed for beating this $d^\ell$ running
time~\cite{Ivakhnenko71,SSM92,KST09,APVZ14,DKKS14}, all
but~\cite{SSM92} are batch algorithms (and suffer from similar
drawbacks as boosting).
The method of~\cite{SSM92} uses online optimization together with an
adaptive rule for creating interaction features.
A variant of this is discussed in Section~\ref{sec:algorithm} and is
used in the experimental study in Section~\ref{sec:experiments} as a
baseline.

\section{Adaptive polynomial expansions}
\label{sec:algorithm}

This section describes our new learning algorithm, \alg, which is
based on stochastic gradient descent and explicit feature expansions
that are adaptively defined.  The specific feature expansion
strategy used in \alg is justified in some simple 
examples, and the use of stochastic gradient descent is backed by a
new regret analysis for shifting comparators.

\subsection{Algorithm description}

\begin{algorithm}[t]
  \caption{Adaptive Polynomial Expansion (\alg)}
  \label{alg:main}
  \begin{algorithmic}[1]
    \renewcommand{\algorithmicrequire}{\textbf{input}}

    \REQUIRE
    Initial features $S_1 = \{x_1,\ldots,x_d\}$,
    expansion sizes $(s_k)$,
    epoch schedule $(\tau_k)$,
    stepsizes $(\eta_t)$.

    \STATE Initial weights $\vw_1 := \v0$, initial epoch $k := 1$,
    parent set $\parent_1 := \emptyset$.

    \FOR{$t = 1,2,\dotsc$:}

      \STATE Receive stochastic gradient $\vg_t$.

      \STATE Update weights:
        \quad $\vw_{t+1} := \vw_t - \eta_t [\vg_t]_{S_k}$, \\
      where $[\cdot]_{S_k}$ denotes restriction to monomials in the
      feature set $S_k$.

      \IF{$t = \tau_k$}

      \STATE Let $M_k \subseteq S_k$ be the top $s_k$ monomials
      $m(\vx) \in S_k$ such that $m(\vx) \notin \parent_k$, ordered
      from highest-to-lowest by the weight magnitude in $\vw_{t+1}$.
      \label{step:select}

        \STATE Expand feature set:
          \quad
          $S_{k+1} := S_k \cup \{ x_i \cdot m(\vx) : i \in [d], m(\vx)
          \in M_k \}$, \quad \mbox{and}\\ 
          \qquad\qquad\qquad\qquad\quad
          $\parent_{k+1} := \parent_k \cup \{m(\vx) : m(\vx) \in M_k \}$.
          \label{step:expand}

        \STATE $k := k + 1$.

      \ENDIF
    \ENDFOR

  \end{algorithmic}
\end{algorithm}

The pseudocode for \alg is given in Algorithm~\ref{alg:main}.  We
regard weight vectors $\vw_t$ and gradients $\vg_t$ as members of a
vector space with coordinate basis corresponding to monomials over the
base variables $\vx = (x_1, x_2, \dotsc, x_d)$, up to some large but
finite maximum degree.  

The algorithm proceeds as stochastic gradient descent over the current
feature set to update a weight vector.  At specified times $\tau_k$,
the feature set $S_k$ is expanded to $S_{k+1}$ by taking the top
monomials in the current feature set, ordered by weight magnitude in
the current weight vector, and creating interaction features between
these monomials and $\vx$. Care is exercised to not
repeatedly pick the same monomial for creating higher order monomial
by tracking a parent set $\parent_k$, the set of all monomials for
which higher degree terms have been expanded.
We provide more
intuition for our choice of this feature growing heuristic in
Section~\ref{sec:heuristics}.

There are two benefits to this staged process.
Computationally, the stages allow us to amortize the cost of the
adding of monomials---which is implemented as an expensive dense
operation---over several other (possibly sparse) operations.
Statistically, using
stages guarantees that the monomials added in the previous stage have
an opportunity to have their corresponding parameters converge.
%
We have found it empirically effective to set $s_k := \operatorname{average}
\norm{[\vg_t]_{S_1}}_0$, and to update the feature set at a constant
number of equally-spaced times over the entire course of learning.  In
this case, the number of updates (plus one) bounds the maximum degree
of any monomial in the final feature set.

\subsection{Shifting comparators and a regret bound for regularized objectives}
\label{sec:regret}

Standard regret bounds compare the cumulative loss of an online
learner to the cumulative loss of a \emph{single} predictor (comparator) from
a fixed comparison class.  \emph{Shifting regret} is a more general
notion of regret, where the learner is compared to a
\emph{sequence} of comparators $\vu_1, \vu_2, \dotsc, \vu_T$.

Existing shifting regret bounds can be used to loosely
justify the use of online gradient descent methods over expanding
feature spaces~\cite{Zin03}.
These bounds are roughly of the form
$\sum_{t=1}^T f_t(\vw_t) - f_t(\vu_t) \lesssim \sqrt{T \sum_{t<T} \norm{\vu_t - \vu_{t+1}}}$,
where $\vu_t$ is allowed
to use the same features available to $\vw_t$, and $f_t$ is the convex cost function in step $t$.
This suggests a relatively high cost for 
a substantial total change in the comparator, and thus in the feature space.
Given a budget, one could either do a liberal expansion a small number of 
times, or opt for including a small number of carefully chosen
monomials more frequently.
We have found that the
computational cost of carefully picking a small number 
of high quality monomials is often quite high.
With computational considerations at the forefront,
we will prefer a more liberal but infrequent expansion.
This also effectively exposes the
learning algorithm to a large number of nonlinearities quickly,
allowing their parameters to jointly converge between the stages.

It is natural to ask if better guarantees are possible under some
structure on the learning problem. Here, we consider the
stochastic setting (rather than the harsher adversarial setting
of~\cite{Zin03}), and further assume that our objective takes the
form
\begin{equation}
  f(\vw) := \bbE[\ell(\ip{\vw}{\vx y})] + \lambda\|\vw\|^2/2,
  \label{eqn:reg-linear}
\end{equation}
where the expectation is under the (unknown) data generating
distribution $D$ over $(\vx,y) \in S \times \bbR$, and $\ell$ is some
convex loss function on which suitable restrictions will be
placed. Here $S$ is such that $S_1 \subseteq S_2 \subseteq \ldots
\subseteq S$, based on the largest degree monomials we intend to
expand. We assume that in round $t$, we observe a stochastic gradient
of the objective $f$, which is typically done by first sampling
$(\vx_t, y_t) \sim D$ and then evaluating the gradient of the
regularized objective on this sample.

This setting has some interesting
structural implications over the general setting of online learning
with shifting comparators.
%
First, the fixed objective $f$ gives us a more direct way of tracking
the change in comparator through $f(\vu_{t}) - f(\vu_{t+1})$,
which might often be milder than $\norm{\vu_{t} - \vu_{t+1}}$.
In particular, if $\vu_t = \arg\min_{\vu \in S_k} f(\vu)$ in epoch $k$,
for a nested subspace sequence $S_k$, then we immediately obtain
$f(\vu_{t+1}) \leq f(\vu_t)$.
Second, the strong convexity of the regularized objective enables the
possibility of faster $O(1/T)$ rates than prior work~\cite{Zin03}.
Indeed, in this setting, we obtain the following stronger
result.
We use the
shorthand $\bbE_t[\cdot]$ to denote the conditional expectation at
time $t$, conditioning over 
the data from rounds $1,\ldots,t-1$.

\begin{theorem}
    \label{fact:tracking:fancier}
    Let a distribution over $(\vx,y)$, twice differentiable convex
    loss $\ell$ with $\ell \geq 0 $ and $\max\{\ell',\ell''\}\leq 1$,
    and a regularization parameter $\lambda > 0$ be given. Recall the
    definition~\eqref{eqn:reg-linear} of the objective $f$. Let
    $(\vw_t, \vg_t)_{t\geq 1}$ be as specified by \alg with step size
    $\eta_t := 1/(\lambda(t+1))$,
    where $\bbE_t([\vg_t]_{\FS{t}}) = [\nf(\vw_t)]_{\FS{t}}$
    and $\FS{t}$ is the support set corresponding to
    epoch $k_t$ at time $t$ in \alg.
    Then for any comparator sequence
    $(\vu_t)_{t=1}^\infty$ satisfying $\vu_t \in \FS{t}$, for any fixed
    $T\geq 1$,
    \begin{align*}
        \bbE\left(
            f(\vw_{T+1})
            -
            \frac{\sum_{t=1}^{T} (t+2)f(\vu_{t})}{\sum_{t=1}^T (t+2)}
        \right)
        &\leq \frac {1}{T+1}\left(
        \frac {(X^2+\lambda)(X+\lambda D)^2}{2\lambda^2}
        \right),
    \end{align*}
    where $X\geq \max_t\|\vx_ty_t\|$ and $D \geq \max_t\max\{\|\vw_t\|,\|\vu_t\|\}$.
\end{theorem}

Quite remarkably, the result exhibits no dependence on the cumulative
shifting of the comparators unlike existing bounds~\cite{Zin03}. This
is the first result of this sort amongst shifting bounds to the best
of our knowledge, and the only one that yields $1/T$ rates of
convergence even with strong convexity, something that the standard
analysis fails to do. Of course, we limit ourselves to the stochastic
setting for this improvement, and prove expected regret guarantees on
the final predictor $\vw_T$ as opposed to a bound on
$\sum_{t=1}^Tf(\vw_t)/T$ which is often studied even in stochastic
settings. 

A curious distinction is our comparator, which we believe gives us
intuition for the source of our improved result.  Note that standard
shifting regret bounds~\cite{Zin03} can be thought of as comparing
$f(\vw_t)$ to $f(\vu_t)$, which is a harder comparison than the
weighted average of $f(\vu_t)$ that we compare to---we discuss the
particular non-uniform weighting in the next paragraph.  Critically,
averages are slower moving objects and hence the yardsticks at time
$t$ and $t+1$ differ only by $O(1/t)$. This observation can be
immediately combined with the known results of Zinkevich~\cite{Zin03}
to show a $O(\sqrt{T})$ \emph{cumulative regret} bound against an
averaged comparator sequence, without needing any strong convexity or
smoothness assumptions on $f$. However, it does not immediately
yield rates on the individual iterates $f(\vw_t)$ even after making
these additional assumptions. Given the way our algorithm utilizes the
weights to grow the support sets, such a guarantee is necessary and
hence establish the result in Theorem~\ref{fact:tracking:fancier}.

As mentioned above,
our comparator is a \emph{weighted average} of $f(\vu_t)$ as
opposed to the more standard uniform average.
Supposing again that $f(\vu_{t+1}) \leq f(\vu_t)$, the weighted average comparator
is a \emph{strictly
  harder benchmark} than an unweighted average and overemphasizes the
later comparator terms which are based on larger support sets. Indeed,
this is a nice compromise between competing against $\vu_T$, which is
the hardest yardstick, and $\vu_1$, which is what a standard
non-shifting analysis compares to. Overall, this result demonstrates
that in our setting, while there is generally a cost to be paid for
shifting the comparator too much, it can still be effectively
controlled in favorable cases. One problem for future work is to
establish these fast $1/T$ rates also with high probability;
as detailed in \Cref{sec:proofs} (which moreover contains the proof of \Cref{fact:tracking:fancier}),
existing techniques yield only an $O(\sqrt{T})$ bound on the deviation
term.

\subsection{Feature expansion heuristics}
\label{sec:heuristics}

Previous work on learning sparse polynomials~\cite{SSM92} suggests
that it is possible to anticipate the utility of interaction features
before even evaluating them.  For instance, one of the algorithms
from~\cite{SSM92} orders monomials $m(\vx)$ by an estimate of $\bbE[
  r(\vx)^2 m(\vx)^2 ] / \bbE[ m(\vx)^2 ]$,
where $r(\vx) = \bbE[y|\vx] - \hat{f}(\vx)$ is the residual of the
current predictor $\hat{f}$ (for least-squares prediction of the label
$y$).  Such an index is shown to be related to the potential error
reduction by polynomials with $m(\vx)$ as a factor.
We call this the SSM heuristic
(after the authors of~\cite{SSM92},
though it differs from their original algorithm).

Another plausible heuristic, which we use in
Algorithm~\ref{alg:main}, simply orders the monomials in $S_k$ by
their weight magnitude in the current weight vector.
We can justify this weight heuristic in the following simple example.
Suppose a target function $\bbE[y|\vx]$ is just a single monomial in
$\vx$, say, $m(\vx) := \prod_{i \in M} x_i$ for some $M \subseteq
[d]$, and that $\vx$ has a product distribution over $\{0,1\}^d$ with
$0 < \bbE[x_i] =: p \leq 1/2$ for all $i \in [d]$.  Suppose we
repeatedly perform $1$-sparse regression with the current feature set
$S_k$, and pick the top weight magnitude monomial for inclusion in the
parent set $\parent_{k+1}$. It is easy to show that the weight on a
degree $\ell$ sub-monomial of $m(\vx)$ in this regression is $p^{|M| -
  \ell}$, and the weight is strictly smaller for any term which is not a
proper sub-monomial of $m(\vx)$. Thus we repeatedly pick the largest
available sub-monomial of $m(\vx)$ and expand it,
eventually discovering $m(\vx)$. After $k$ stages of the algorithm, we
have at most $kd$ features in our regression here, and hence we find
$m(\vx)$ with a total of $d|M|$ variables in our regression, as
opposed to $d^{|M|}$ which typical feature selection approaches would
need. This intuition can be extended more generally to scenarios where
we do not necessarily do a sparse regression and beyond product
distributions, but we find that even this simplest example illustrates
the basic motivations underlying our choice---we want to
parsimoniously expand on top of a base feature set, while still
making progress towards a good polynomial for our data. 

\subsection{Fall-back risk-consistency}

Neither the SSM heuristic nor the weight heuristic is rigorously
analyzed (in any generality).  Despite this, the basic algorithm \alg
can be easily modified to guarantee a form of risk consistency,
regardless of which feature expansion heuristic is used. Consider the
following variant of the support update rule in the algorithm
\alg. Given the current feature budget $s_k$, we add $s_{k}-1$
monomials ordered by weight magnitudes as in
Step~\ref{step:expand}. We also pick a monomial $m(\vx)$ of the
smallest degree such that $m(\vx) \notin \parent_k$. Intuitively, this
ensures that all degree 1 terms are in $\parent_k$ after $d$ stages,
all degree 2 terms are in $\parent_k$ after $k = O(d^2)$ stages and so
on. In general, it is easily seen that $k = O(d^{\ell-1})$ ensures
that all degree $\ell-1$ monomials are in $\parent_k$ and hence all
degree $\ell$ monomials are in $S_k$. For ease of exposition, let us
assume that $s_k$ is set to be a constant $s$ independent of $k$. Then
the total number of monomials in $\parent_k$ when $k = O(d^{\ell-1})$
is $O(sd^{\ell-1})$, which means the total number of features in $S_k$
is $O(sd^\ell)$.

Suppose we were interested in competing with all $\spindex$-sparse
polynomials of degree $\ell$. The most direct approach would be to
consider the explicit enumeration of all monomials of degree up
to $\ell$, and then perform $\ell_1$-regularized
regression~\cite{Tibshirani96b} or a greedy variable selection method
such as OMP~\cite{TroppGil07} as means of enforcing sparsity. This
ensures consistent estimation with $n = O(\spindex\log d^{\ell}) =
O(\spindex\ell\log d)$ examples. In contrast, we might need $n =
O(\spindex(\ell\log d + \log s))$ examples in the worst case using
this fall back rule, a minor overhead at best. However, in favorable
cases, we stand to gain a lot when the heuristic succeeds in finding
good monomials rapidly. Since this is really an empirical question, we
will address it with our empirical evaluation.

\section{Experimental study}
\label{sec:experiments}

We now describe our empirical evaluation of \alg.

\subsection{Implementation, experimental setup, and performance
metrics}

In order to assess the effectiveness of our algorithm, it is critical
to build on top of an efficient learning framework that can handle
large, high-dimensional datasets. To this end, we implemented \alg in
the Vowpal Wabbit (henceforth VW) open source machine learning
software\footnote{%
    Please see
   \url{https://github.com/JohnLangford/vowpal\_wabbit}
   and the associated git repository, where
   \texttt{--stage\_poly} and related command line options
   execute \alg.%
}.
VW is a good framework
for us, since it also natively supports quadratic and cubic expansions
on top of the base features.
These expansions are done dynamically at run-time, rather than being
stored and read from disk in the expanded form for computational
considerations. To deal with these dynamically enumerated features, VW
uses hashing to associate features with indices, mapping each feature
to a $b$-bit index, where $b$ is a parameter. The core learning
algorithm is an online algorithm as assumed in \alg, but uses
refinements of the basic stochastic gradient descent update
(e.g.,~\cite{DuchiHS2011,McMahanS2010,KarampatziakisL2011,RossML2013}).

We implemented \alg such that the total number of epochs was always 6
(meaning 5 rounds of adding new features). At the end of each epoch,
the non-parent monomials with largest magnitude weights were marked as
parents. Recall that the number of parents is modulated at $s^\alpha$
for some $\alpha > 0$, with $s$ being the average number of non-zero
features per example in the dataset so far. We will present
experimental results with different choices of $\alpha$, and we found
$\alpha = 1$ to be a reliable default. Upon seeing an example,
the features are enumerated on-the-fly by recursively expanding the
marked parents, taking products with base monomials. These operations
are done in a way to respect the sparsity (in terms of base
features) of examples which many of our datasets exhibit.

Since the benefits of nonlinear learning over linear learning
themselves are very dataset dependent, and furthermore can vary
greatly for different heuristics based on the problem at hand, we
found it important to experiment with a large testbed consisting of a
diverse collection of medium and large-scale datasets. To this end, we
compiled a collection of 30 publicly available datasets, across a
number of KDDCup challenges, UCI repository and other common resources
(detailed in the appendix). For all the datasets, we tuned the
learning rate for each learning algorithm based on the
progressive validation error (which is typically a reliable bound on
test error)~\cite{BlumKL1999}. The number of bits in hashing was set
to 18 for all algorithms, apart from cubic polynomials, where using 24
bits for hashing was found to be important for good statistical
performance. For each dataset, we performed a random split with 80\%
of the data used for training and the remainder for testing. For all
datasets, we used squared-loss to train, and $0$-$1$/squared-loss for
evaluation in classification/regression problems. We also experimented
with $\ell_1$ and $\ell_2$ regularization, but these did not help
much. The remaining settings were left to their VW defaults.

\begin{figure*}[t]
  \centering
  \begin{tabular}{@{}c@{}c@{}}
    \includegraphics[width=0.5\textwidth]{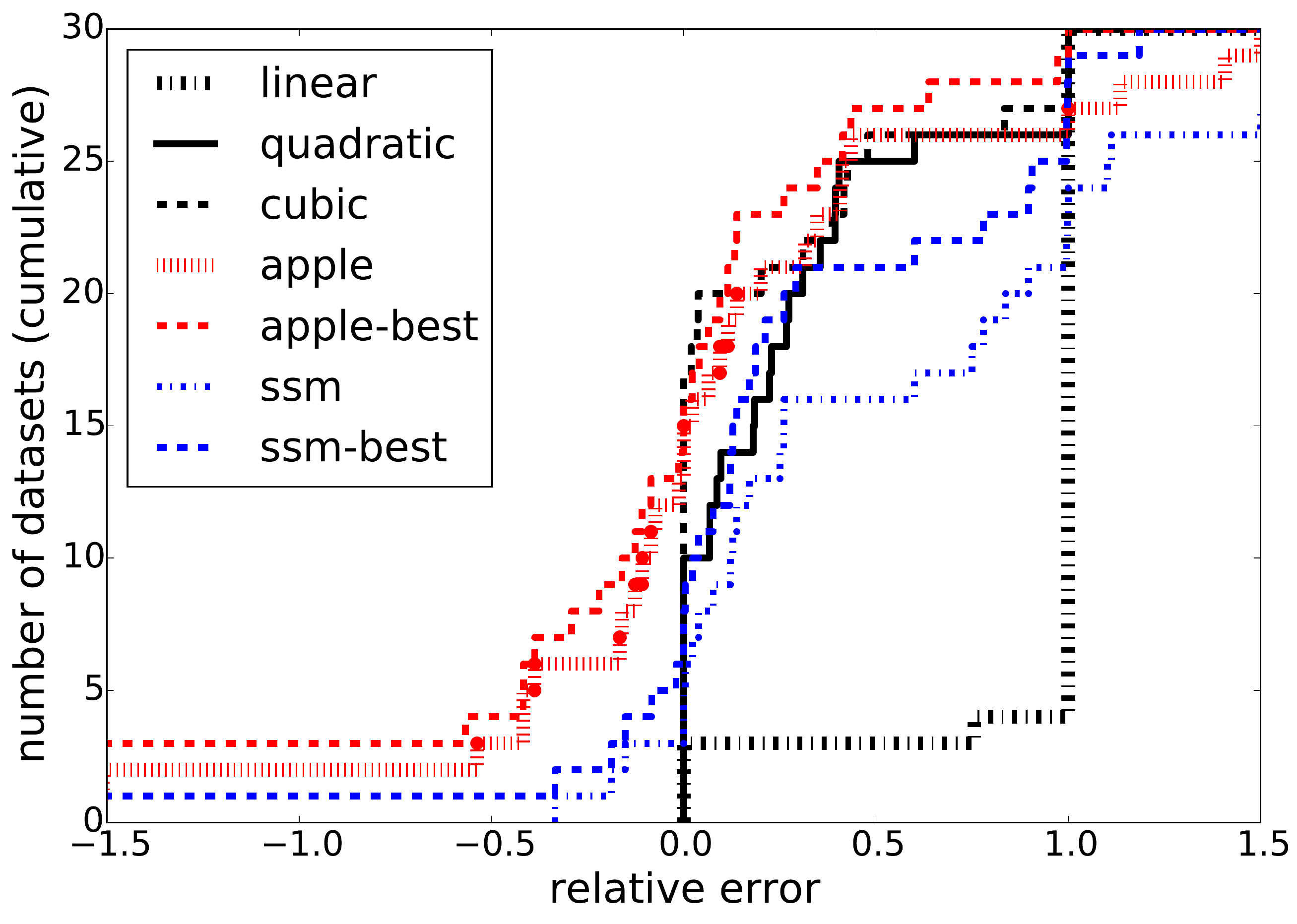} & 
    \includegraphics[width=0.5\textwidth]{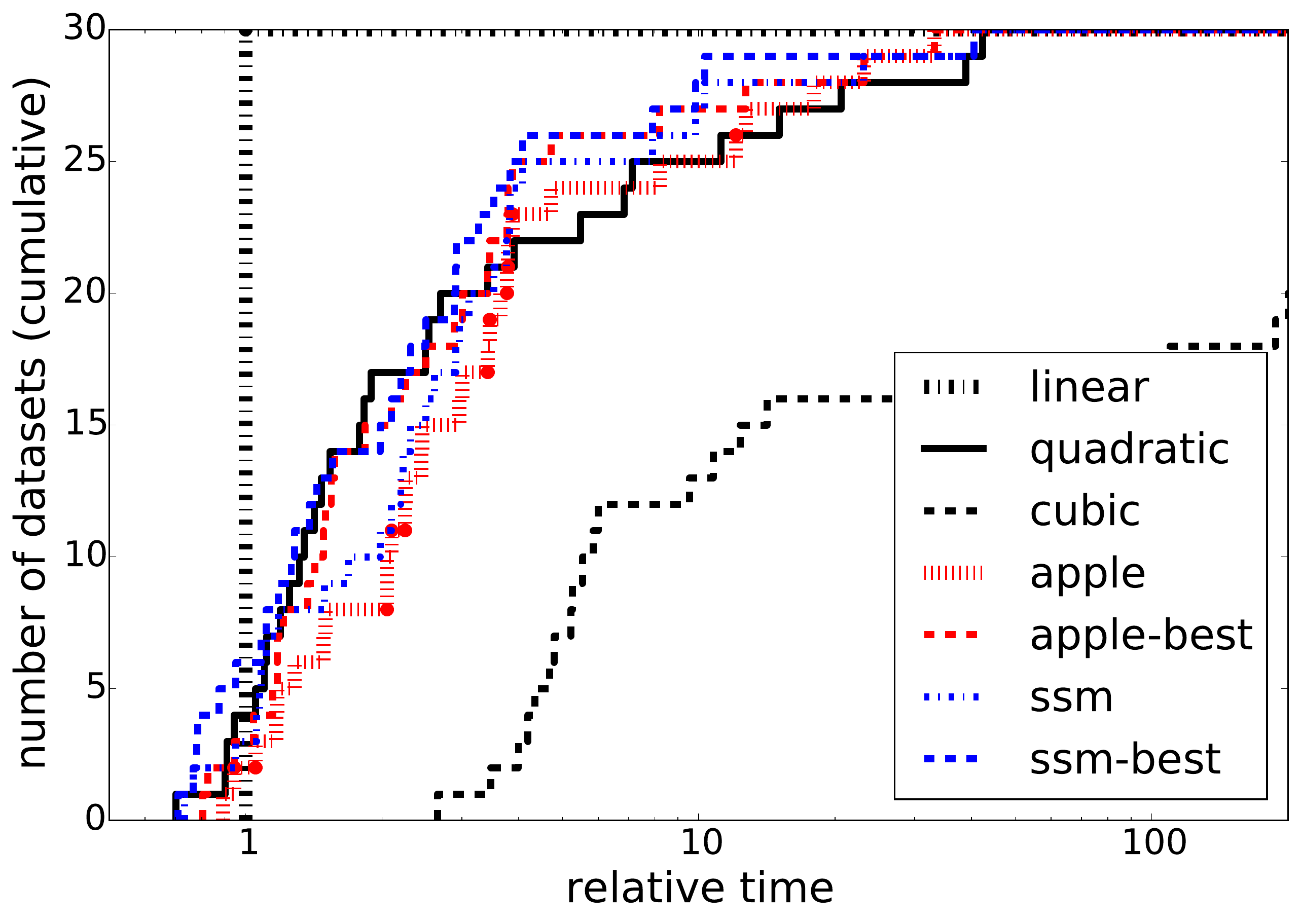}\\
    (a) & (b)
  \end{tabular}
  \caption{%
    Dataset CDFs across all 30 datasets: (a) relative test error, (b)
    relative training time (log scale). $\{\alg, \ssm\}$ refer to the
    $\alpha = 1$ default; $\{\alg,\ssm\}$-best picks best $\alpha$ per
    dataset.%
  }
  \label{fig:cdfs-full}
\end{figure*}

For aggregating performance across 30 diverse datasets, it was
important to use error and running time measures on a scale
independent of the dataset. Let $\linear$, $\quadratic$ and $\cubic$
refer to the test errors of linear, quadratic and cubic baselines
respectively (with \lalg, \qalg, and \calg used to denote the
baseline algorithms themselves). For an algorithm $\algplain$, we compute the
\emph{relative (test) error}:
\begin{equation}
  \rerr(\algplain) = \frac{\mathrm{err}(\algplain) - \min(\linear, \quadratic,
    \cubic)}{\max(\linear, \quadratic, \cubic) - \min(\linear, \quadratic,
    \cubic)}, 
  \label{eq:rel-error}
\end{equation}
where $\min(\linear, \quadratic, \cubic)$ is the smallest error
among the three baselines on the dataset, and
$\max(\linear, \quadratic, \cubic)$ is similarly defined. We also
define the \emph{relative (training)
  time} as the ratio to running time of $\lalg$:
  $\rtime(\algplain) = \runtime(\algplain)/\runtime(\lalg)$.
  With these definitions, the aggregated
plots of relative errors and relative times for the various baselines
and our methods are shown in Figure~\ref{fig:cdfs-full}. For each
method, the plots show a cumulative distribution function (CDF) across
datasets: an entry $(a,b)$ on the left plot indicates that
the relative error for $b$ datasets was at most $a$. The plots include
the baselines $\lalg, \qalg, \calg$, as well as a variant of \alg
(called \ssm) that
replaces the weight heuristic with the SSM heuristic,
as described in
Section~\ref{sec:heuristics}.
For \alg and \ssm,
the plot shows the results with the fixed setting of $\alpha = 1$, as
well as the best setting chosen per dataset from $\alpha \in \{0.125,
0.25, 0.5, 0.75, 1\}$ (referred to as \alg-best and \ssm-best). 

\subsection{Results}

In this section, we present some aggregate results. Detailed results
with full plots and tables are presented in the appendix. In the
Figure~\ref{fig:cdfs-full}(a), the relative error for all of \lalg,
\qalg and \calg is always to the right of 0 (due to the definition of
$\rerr$). In this plot, a curve enclosing a larger area indicates, in
some sense, that one method uniformly dominates another. Since \alg
uniformly dominates \ssm statistically (with only slightly longer
running times), we restrict the remainder of our study to comparing
\alg to the baselines \lalg, \qalg and \calg. We found that on 12 of
the 30 datasets, the relative error was negative, meaning that \alg
beats all the baselines. A relative error of 0.5 indicates that we
cover at least half the gap between $\min(\linear, \quadratic,
\cubic)$ and $\max(\linear, \quadratic, \cubic)$. We find that we are
below 0.5 on 27 out of 30 datasets for \alg-best, and 26 out of the 30
datasets for the setting $\alpha = 1$. This is particularly striking
since the error $\min(\linear, \quadratic, \cubic)$ is attained by
\calg on a majority of the datasets (17 out of 30), where the relative
error of \calg is 0. Hence, statistically \alg often outperforms even
\calg, while typically using a much smaller number of features.  To
support this claim, we include in the appendix a plot of the average
number of features per example generated by each method, for all
datasets. Overall, we find the statistical performance of \alg from
Figure~\ref{fig:cdfs-full} to be quite encouraging across this large
collection of diverse datasets.

\begin{figure*}[t]
  \centering
  \begin{tabular}{@{}c@{}c@{}}
    \includegraphics[width=0.5\textwidth]{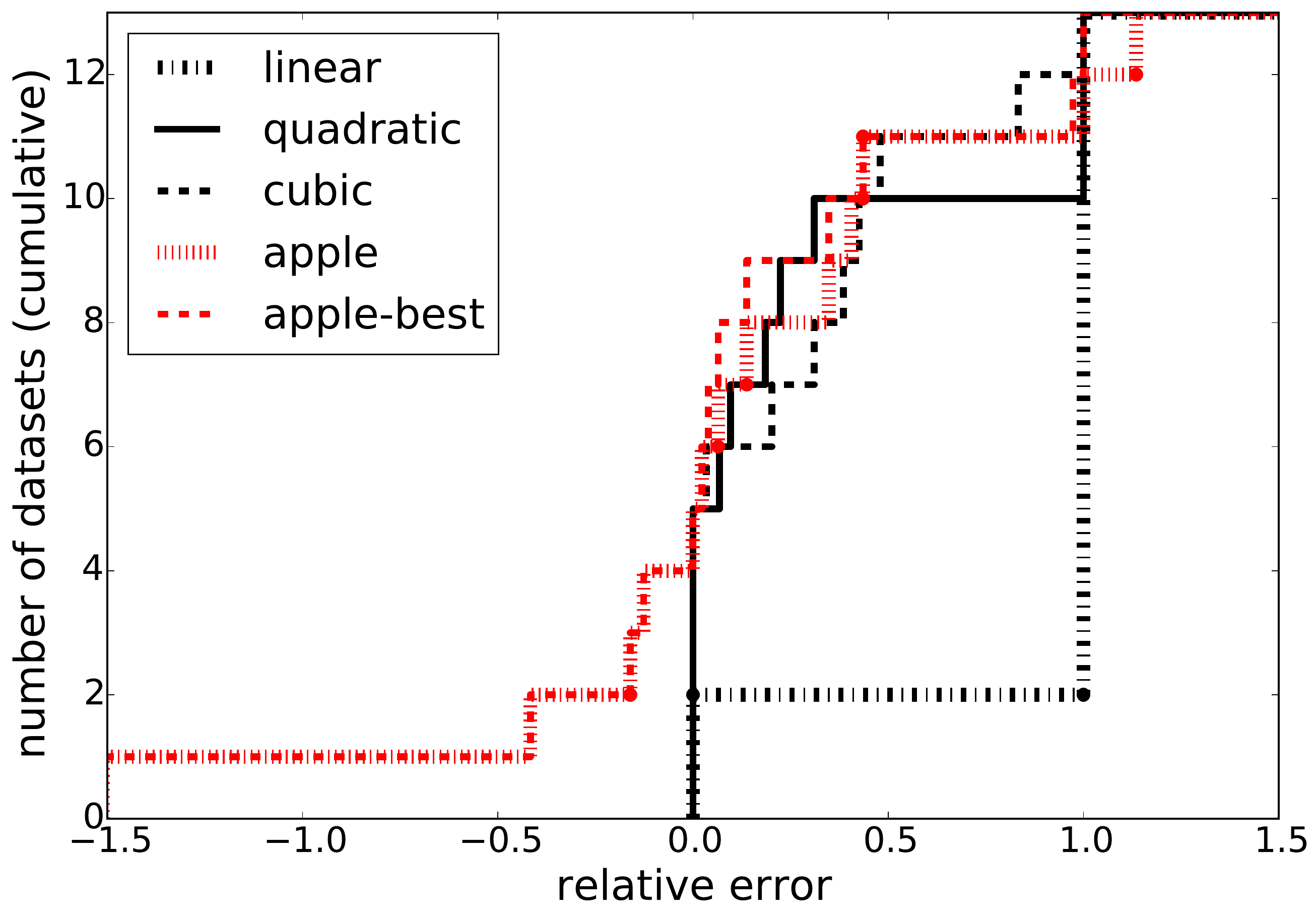} & 
    \includegraphics[width=0.5\textwidth]{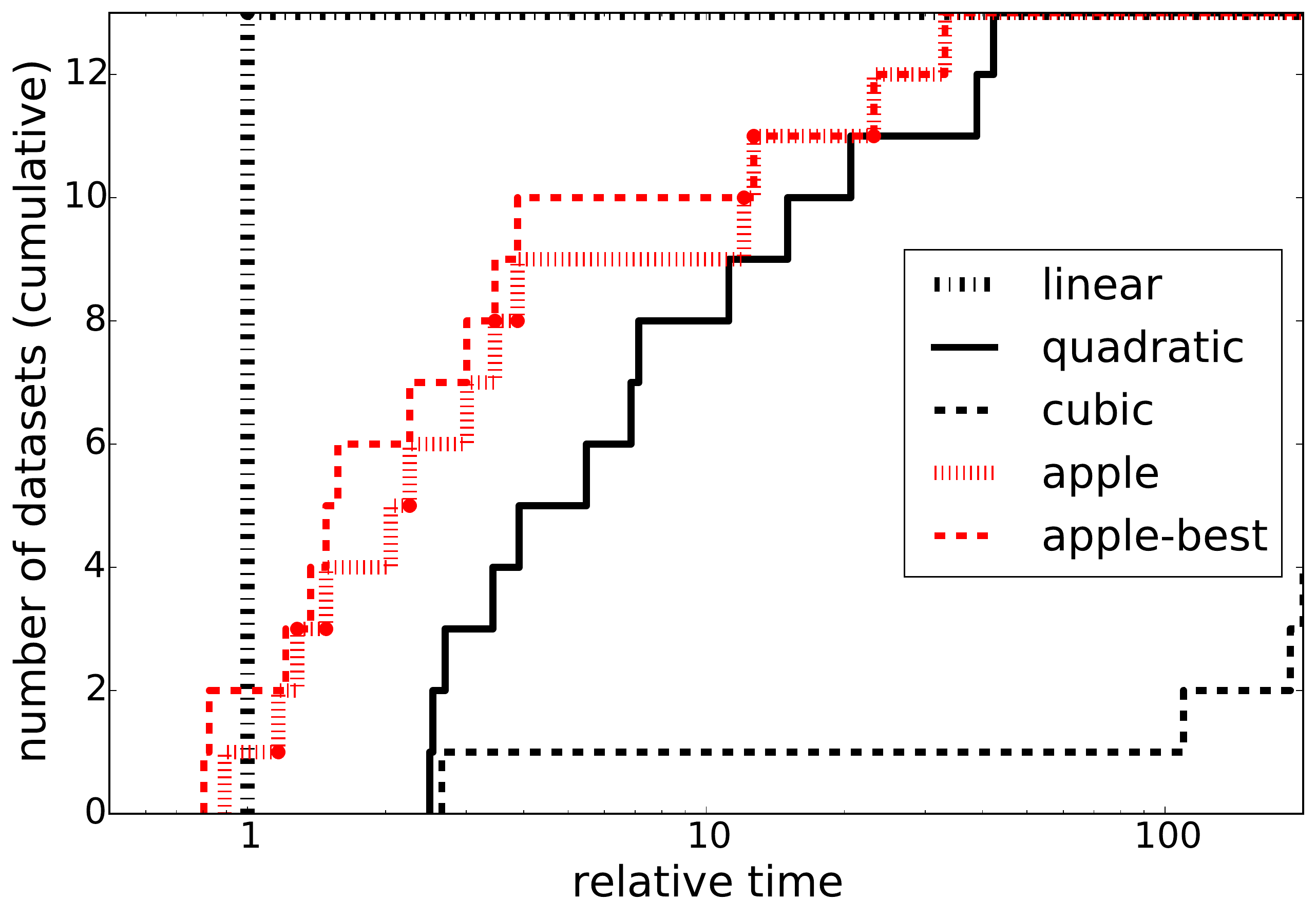}\\
    (a) & (b)
  \end{tabular}
  \caption{%
    Dataset CDFs across 13 datasets where $\runtime(\qalg) \geq
    2 \runtime(\lalg)$:
    (a) relative test error,
    (b) relative training time (log scale).%
  }
  \vspace{-0.5cm}
  \label{fig:cdfs-filt}
\end{figure*}

The running time performance of \alg is also extremely good.
Figure~\ref{fig:cdfs-full}(b) shows that the running time of
\alg is within a factor of 10 of \lalg for almost all datasets, which
is quite impressive considering that we generate a potentially much
larger number of features. The gap between \lalg and \alg is
particularly small for several large datasets, where the examples are
sparse and high-dimensional. In these cases, all algorithms are
typically I/O-bottlenecked, which is the same for all algorithms due
to the dynamic feature expansions used. It is easily seen that the
statistically efficient baseline of \calg is typically computationally
infeasible, with the relative time often being as large as $10^2$ and
$10^5$ on the biggest dataset. Overall,
the statistical performance of \alg is competitive with and
often better than $\min(\linear, \quadratic, \cubic)$, and offers
a nice intermediate in computational complexity.

A surprise in Figure~\ref{fig:cdfs-full}(b) is that \qalg
appears to computationally outperform \alg for a relatively large
number of datasets, at least in aggregate.
This is due to the extremely efficient implementation of \qalg in VW: on 17 of 30 datasets, the
running time of \qalg is less than twice that of \lalg. While we often
statistically outperform \qalg on many of these smaller datasets, we
are primarily interested in
the larger datasets where the relative
cost of nonlinear expansions (as in \qalg) is high.

In Figure~\ref{fig:cdfs-filt}, we restrict attention to the 13
datasets where $\runtime(\qalg) / \runtime(\lalg) \geq 2$.  On these
larger datasets, our statistical performance seems to dominate all the
baselines (at least in terms of the CDFs, more on individual datasets
will be said later). In terms of computational time, we see that we
are often much better than \qalg, and \calg is essentially infeasible
on most of these datasets. This demonstrates our key intuition that
such adaptively chosen monomials are key to effective nonlinear
learning in large, high-dimensional datasets.

We also experimented with \emph{picky} algorithms of the sort
mentioned in Section~\ref{sec:regret}.  We tried the original
algorithm from~\cite{SSM92}, which tests a candidate monomial before
adding it to the feature set $S_k$, rather than just testing candidate
parent monomials for inclusion in $\parent_k$; and also a picky
algorithm based on our weight heuristic.  Both algorithms were
extremely computationally expensive, even when implemented using VW as
a base: the explicit testing for inclusion in $S_k$ (on a per-example
basis) caused too much overhead.  We ruled out other baselines such as
polynomial kernels for similar computational reasons.

\begin{figure*}[t]
  \centering
  \begin{tabular}{@{}c@{}c@{}}
    \includegraphics[width=0.5\textwidth]{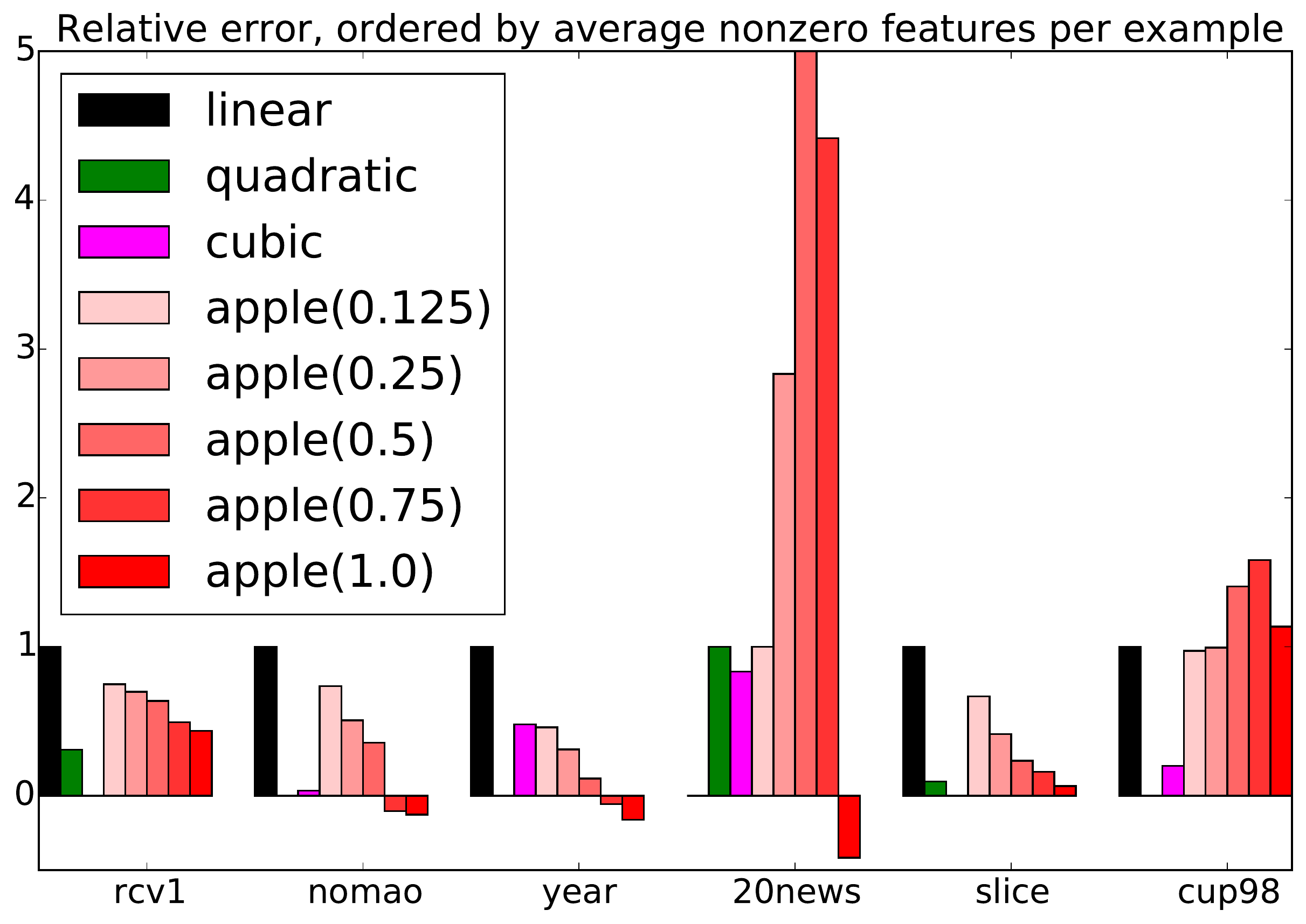} & 
    \includegraphics[width=0.5\textwidth]{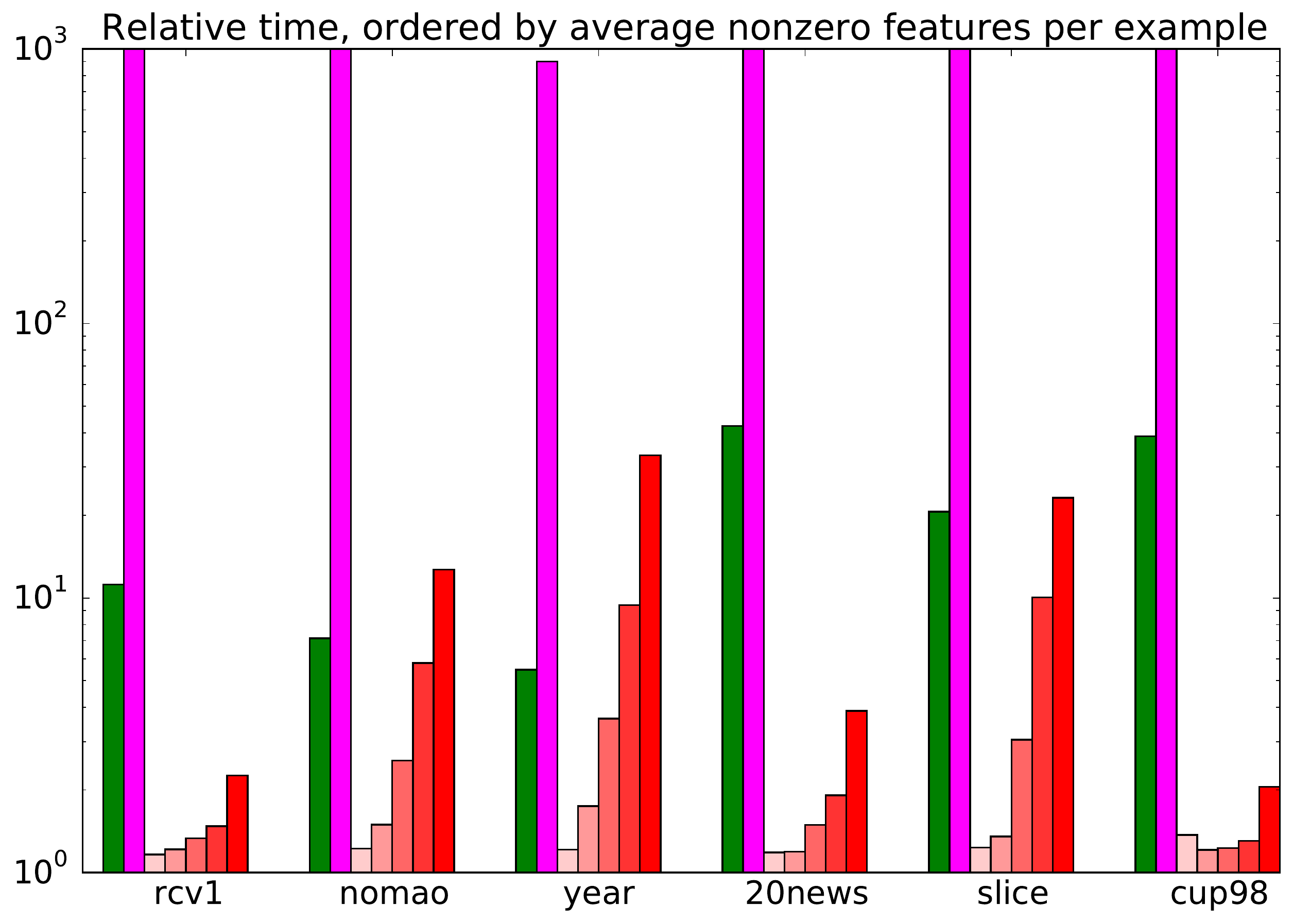}\\
    (a) & (b)
  \end{tabular}
  \caption{%
    Comparison of different methods on the top 6 datasets by non-zero
    features per example:
    (a) relative test errors,
    (b) relative training times.%
  }
  \label{fig:hists-topfeats}
\end{figure*}

To provide more intuition, we also show individual results for
the top 6 datasets with the highest average number of non-zero
features per example---a key factor determining the
computational cost of all approaches. In
Figure~\ref{fig:hists-topfeats}, we show the performance of the \lalg,
\qalg, \calg baselines, as well as \alg with 5 different parameter
settings in terms of relative error
(Figure~\ref{fig:hists-topfeats}(a)) and relative time
(Figure~\ref{fig:hists-topfeats}(b)). The results are overall quite
positive. We see that on 3 of the datasets, we improve upon all the
baselines statistically, and even on other 3 the performance is quite
close to the best of the baselines with the exception of the
\texttt{cup98} dataset. In terms of running time, we find \calg to be
extremely expensive in all the cases. We are typically faster than
\qalg, and in the few cases where we take longer, we also obtain
a statistical improvement for the slight increase in computational
cost. On larger datasets, the performance
of our method is quite desirable and in line with our
expectations.

Finally, we also implemented a parallel version of our algorithm,
building on the repeated averaging approach~\cite{HallGiMa2010,
ACDL14}, using the built-in AllReduce communication mechanism of VW,
and ran an experiment using an
internal advertising dataset consisting of
approximately 690M training examples, with roughly 318 non-zero
features per example. The task is the prediction of
\texttt{click/no-click} events. The data was stored in a large Hadoop
cluster, split over 100 partitions. We implemented the \lalg baseline,
using 5 passes of online learning with repeated averaging on this
dataset, but could not run full \qalg or \calg baselines due to the
prohibitive computational cost. As an intermediate, we generated
$\mathtt{bigram}$ features, which only doubles the number of non-zero
features per example.
We parallelized
\alg as
follows. In the first pass over the data, each one of the 100 nodes
locally selects the promising features over 6 epochs, as in our
single-machine setting. We then take the union of all the parents
locally found across all nodes, and freeze that to be the parent set
for the rest of training. The remaining 4 passes are now done with
this fixed feature set, repeatedly averaging local weights. We then
ran \alg, on top of both \lalg as well as $\mathtt{bigram}$ as the base
features to obtain maximally expressive features. The test error was
measured in terms of the area under ROC curve (AUC), since this is a
highly imbalanced dataset. The error and time results, reported in
Table~\ref{tbl:parallel}, show that using nonlinear features does
lead to non-trivial improvements in AUC, albeit at an increased
computational cost. Once again, this should be put in perspective with
the full \qalg baseline, which did not finish in over a day on this
dataset.

\begin{table}
  \centering
  \begin{tabular}{|c|c|c|c|c|}
    \hline
    ~&$\lalg$&$\lalg+\alg$&$\balg$&$\balg+\alg$\\ \hline 
    Test AUC & 0.81664  & 0.81712 & 0.81757 & 0.81796\\ \hline
    Training time (in s) & 1282 & 2727 & 2755 & 7378\\ \hline
  \end{tabular}
  \caption{Test error and training times for different methods in a
    large-scale distributed setting. For $\{\lalg,\balg\}+\alg$, we
    used $\alpha = 0.25$.}
  \label{tbl:parallel}
\end{table}


\paragraph{Acknowledgements:} The authors would like to thank Leon
Bottou, Rob Schapire and Dean Foster who were involved in several
formative and helpful discussions. 

\bibliography{paper} \bibliographystyle{plain}

\begin{thebibliography}{10}

\bibitem{ACDL14}
Alekh Agarwal, Olivier Chapelle, Miroslav Dud{\'i}k, and John Langford.
\newblock A reliable effective terascale linear learning system.
\newblock {\em Journal of Machine Learning Research}, 15(Mar):1111--1133, 2014.

\bibitem{APVZ14}
Alexandr Andoni, Rina Panigrahy, Gregory Valiant, and Li~Zhang.
\newblock Learning sparse polynomial functions.
\newblock In {\em SODA}, 2014.

\bibitem{BlumKL1999}
Avrim Blum, Adam Kalai, and John Langford.
\newblock Beating the hold-out: Bounds for k-fold and progressive
  cross-validation.
\newblock In {\em COLT}, 1999.

\bibitem{BordesEWB2005}
Antoine Bordes, Seyda Ertekin, Jason Weston, and L\'{e}on Bottou.
\newblock Fast kernel classifiers with online and active learning.
\newblock {\em Journal of Machine Learning Research}, 6:1579--1619, 2005.

\bibitem{Bubeck14}
S{\'e}bastien Bubeck.
\newblock Theory of convex optimization for machine learning.
\newblock 2014.
\newblock {\tt arXiv:1405.4980 [math.OC]}.

\bibitem{DKKS14}
Alexandros~G. Dimakis, Adam Klivans, Murat Kocaoglu, and Karthikeyan Shanmugam.
\newblock A smoothed analysis for learning sparse polynomials.
\newblock {\em CoRR}, abs/1402.3902, 2014.

\bibitem{DuchiHS2011}
John Duchi, Elad Hazan, and Yoram Singer.
\newblock Adaptive subgradient methods for online learning and stochastic
  optimization.
\newblock {\em The Journal of Machine Learning Research}, 12:2121--2159, 2011.

\bibitem{FS97}
Yoav Freund and Robert~E. Schapire.
\newblock A decision-theoretic generalization of on-line learning and an
  application to boosting.
\newblock {\em Journal of Computer and System Sciences}, 55(1):119--139, 1997.

\bibitem{Friedman99}
Jerome~H. Friedman.
\newblock Greedy function approximation: A gradient boosting machine.
\newblock Technical report, Department of Statistics, Stanford University,
  1999.

\bibitem{HallGiMa2010}
K.~Hall, S.~Gilpin, and G.~Mann.
\newblock Mapreduce/bigtable for distributed optimization.
\newblock In {\em Workshop on Learning on Cores, Clusters, and Clouds}, 2010.

\bibitem{HGXD14}
Raffay Hamid, Alex Gittens, Ying Xiao, and Dennis Decoste.
\newblock Compact random feature maps.
\newblock In {\em ICML}, 2014.

\bibitem{Ivakhnenko71}
A.~G. Ivakhnenko.
\newblock Polynomial theory of complex systems.
\newblock {\em Systems, Man and Cybernetics, IEEE Transactions on},
  SMC-1(4):364--378, 1971.

\bibitem{JZ14}
Rie Johnson and Tong Zhang.
\newblock Learning nonlinear functions using regularized greedy forest.
\newblock {\em Pattern Analysis and Machine Intelligence, IEEE Transactions
  on}, 36(5):942--954, May 2014.

\bibitem{KST09}
Adam~Tauman Kalai, Alex Samorodnitsky, and Shang-Hua Teng.
\newblock Learning and smoothed analysis.
\newblock In {\em FOCS}, 2009.

\bibitem{KK12}
Purushottam Kar and Harish Karnick.
\newblock Random feature maps for dot product kernels.
\newblock In {\em AISTATS}, 2012.

\bibitem{KarampatziakisL2011}
Nikos Karampatziakis and John Langford.
\newblock Online importance weight aware updates.
\newblock In {\em UAI}, pages 392--399, 2011.

\bibitem{Mahoney11}
Michael~W. Mahoney.
\newblock Randomized algorithms for matrices and data.
\newblock {\em Foundations and Trends in Machine Learning}, 3(2):123--224,
  2011.

\bibitem{McMahanS2010}
H.~Brendan McMahan and Matthew~J. Streeter.
\newblock Adaptive bound optimization for online convex optimization.
\newblock In {\em COLT}, pages 244--256, 2010.

\bibitem{MCFS13}
Indraneel Mukherjee, Kevin Canini, Rafael Frongillo, and Yoram Singer.
\newblock Parallel boosting with momentum.
\newblock In {\em Proceedings of the European Conference on Machine Learning
  and Principles and Practice of Knowledge Discovery in Databases}, 2013.

\bibitem{PP13}
Ninh Pham and Rasmus Pagh.
\newblock Fast and scalable polynomial kernels via explicit feature maps.
\newblock In {\em Proceedings of the 19th ACM SIGKDD International Conference
  on Knowledge Discovery and Data Mining}, 2013.

\bibitem{RR08}
Ali Rahimi and Benjamin Recht.
\newblock Random features for large-scale kernel machines.
\newblock In {\em Advances in Neural Information Processing Systems 20}, 2008.

\bibitem{RossML2013}
St{\'e}phane Ross, Paul Mineiro, and John Langford.
\newblock Normalized online learning.
\newblock In {\em UAI}, 2013.

\bibitem{SSM92}
Terence~D. Sanger, Richard~S. Sutton, and Christopher~J. Matheus.
\newblock Iterative construction of sparse polynomial approximations.
\newblock In {\em Advances in Neural Information Processing Systems 4}, 1992.

\bibitem{SS02}
Bernhard Sch{\"o}lkopf and Alexander Smola.
\newblock {\em Learning with Kernels}.
\newblock MIT Press, Cambridge, MA, 2002.

\bibitem{ShamirZhang13}
Ohad Shamir and Tong Zhang.
\newblock Stochastic gradient descent for non-smooth optimization: Convergence
  results and optimal averaging schemes.
\newblock In {\em ICML}, 2013.

\bibitem{Tibshirani96b}
R.~Tibshirani.
\newblock Regression shrinkage and selection via the lasso.
\newblock {\em J. Royal. Statist. Soc B.}, 58(1):267--288, 1996.

\bibitem{TroppGil07}
J.~A. Tropp and A.~C. Gilbert.
\newblock Signal recovery from random measurements via orthogonal matching
  pursuit.
\newblock {\em IEEE Transactions on Information Theory}, 53(12):4655--4666,
  December 2007.

\bibitem{WS01}
Christopher Williams and Matthias Seeger.
\newblock Using the {N}ystr\"{o}m method to speed up kernel machines.
\newblock In {\em Advances in Neural Information Processing Systems 13}, 2001.

\bibitem{Zin03}
Martin Zinkevich.
\newblock Online convex programming and generalized infinitesimal gradient
  ascent.
\newblock In {\em ICML}, 2003.

\end{thebibliography}

\appendix

\def\R{\mathbb R}
\newcommand{\VE}[2]{\varepsilon^{(#1)}_{#2}}
\def\dev{\textup{dev}}

\section{Proofs and other technical material}
\label{sec:proofs}

The generic statement behind \Cref{fact:tracking:fancier} is as follows;
note that it neither makes specific requirements upon the form of the features (i.e., they need not be monomials),
nor upon how support set $\FS{t+1}$ is derived from $\FS{t}$ (i.e., it only needs to satisfy
the containment $\FS{t+1} \supseteq \FS{t}$).

\begin{theorem}
    \label{fact:tracking:fancy} Let convex function $f$ be given with
    respective strong convexity and strong smoothness parameters
    $\lambda>0$ and $\beta<\infty$. Let $(\vw_t, \vg_t)_{t\geq 1}$ be
    as specified by \alg with step size $\eta_t := 1/(\lambda(t+1))$,
    where $\bbE_t([\vg_t]_{\FS{t}}) = [\nf(\vw_t)]_{\FS{t}}$,
    and $\FS{t}$ is the support set corresponding to
    epoch $k_t$ at time $t$ in \alg,
    with $\FS{t} \subseteq \FS{t+1}$ and $w_1 \in \FS{0}$.
    Then for any comparator sequence $(\vu_t)_{t=1}^\infty$ satisfying
    $\vu_t \in \FS{t}$, for any fixed $T\geq 1$,
    \begin{align*}
        f(\vw_{T+1})
        -
        \frac{\sum_{t=1}^{T} (t+2)f(\vu_{t})}{\sum_{t=1}^T (t+2)}
        &\leq \frac {1}{T+1}\left(
            \frac {\beta G^2}{2\lambda^2}
        +
        \frac 1 {\lambda}\sum_{t=1}^T\dev_t
    \right),
    \end{align*}
    where $G := \max_{t\leq T} \|\vg_t\|$, and the random variable
    \[
        \dev_t := \left(\frac {t+2}{T+2}\right)
        \ip{[\nf(\vw_t)]_{\FS{t}} - [\vg_t]_{\FS{t}}}{\nf(\vw_t)}
    \]
    satisfies $\bbE(\sum_{t=1}^T\dev_t) = 0$ and, with probability at least $1-\delta$ over
    the draw of $\{\vg_t\}_{t=1}^T$,
    $\sum_{t=1}^T\dev_t \leq 4G^2\sqrt{T\ln(1/\delta)}$.
\end{theorem}

As discussed in the main text, the existence of a single target function $f$ (rather than
a new function each round) suggests its use in tracking the progress of the algorithm;
indeed, the proof directly decreases
$f(\vw_{T+1})-\sum_{t=1}^{T} (t+2)f(\vu_{t}) / \sum_{t=1}^T (t+2)$,
rather than passing through a surrogate such as measuring parameter distance $\|\vw_t - \vu_t\|$.
The invocation of smoothness and strong convexity at the core of the argument
(see the display with \cref{eq:fw_ish})
is similar to the analogous invocation of smoothness and boundedness of the domain in the convergence guarantee
for the Frank-Wolfe method~\cite[Theorem 3.4]{Bubeck14}.
This bound is on the
last iterate, whereas the standard proof scheme for subgradient descent, most naturally stated
for averaged iterates~\cite[Theorem 3.1]{Bubeck14}, requires some work for the last
iterate~\cite[Theorem 1]{ShamirZhang13}; on the other hand, the approach here incurs an extra factor $\beta/\lambda$.

\begin{proof}[Proof of \Cref{fact:tracking:fancy}]
    Let $r_T\in\R$ be a parameter (dependent on $T$) left temporarily
    unspecified, and set the quantities

    \begin{align*}
        \VE{1}{t}
        &:= 2\lambda\left(f(\vu_{t}) - r_T\right),
        &\forall t\geq 1,
        \\
        \VE{2}{t}
        &:= \ip{[\nf(\vw_t)]_{\FS{t}} - [\vg_t]_{\FS{t}}}{\nf(\vw_t)},
        &\forall t\geq 1,
        \\
        c
        &:= \frac {\beta G^2}{2\lambda^2}.
    \end{align*}

    To prove the desired bound, it will first be shown, for any $t\geq
    1$, that

    \begin{equation}
        f(\vw_{t+1}) - r_T 
        \leq \left(\frac {t-1}{t+1}\right)\left(f(\vw_t) - r_T\right)
        + \frac {\eta_t^2 \beta G^2}{2}
        + \eta_t\left(\VE{1}{t} + \VE{2}{t}\right).
        \label{eq:sc:1}
    \end{equation}
    Let $t\geq 1$ be arbitrary, and note by strong convexity,
    for any $\vw$ with $\vw \in \FS{t}$,
    since $\vw_t\in \FS{t-1}\subseteq \FS{t}$ and thus $\vw - \vw_t \in \FS{t}$,
    \begin{align*}
        f(\vw)
        &\geq f(\vw_t) + \ip{\nf(\vw_t)}{\vw-\vw_t} + \frac
        {\lambda\|\vw-\vw_t\|_2^2}{2}  
        \\
        &= f(\vw_t) + \ip{[\nf(\vw_t)]_{\FS{t}}}{\vw-\vw_t} + \frac
        {\lambda\|\vw-\vw_t\|_2^2}{2}. 
    \end{align*}
    The right hand side, as a function of $\vw$,
    is a strongly convex quadratic over $\FS{t}$, minimized at
    $\vw_t - [\nf(\vw_t)]_{\FS{t}}/\lambda$.  Plugging this back in,
    \[
        f(\vw) \geq f(\vw_t) - \frac {\|[\nf(\vw_t)]_{\FS{t}}\|_2^2}{2\lambda},
    \]
    which in particular holds for $\vw = \vu_{t}$
    (which satisfies $\vu_{t}\in \FS{t}$), meaning
    \[
        f(\vu_{t}) \geq f(\vw_t) - \frac {\|[\nf(\vw_t)]_{\FS{t}}\|_2^2}{2\lambda}.
    \]
    Combining this with smoothness and the definition of $\vw_{t+1}$,
    \begin{align}
        f(\vw_{t+1})
        -r_T
        &
        \leq f(\vw_t) - r_T
        + \ip{\nf(\vw_t)}{-\eta_t [\vg_t]_{\FS{t}}}
        + \frac {\eta_t^2 \beta \| [\vg_t]_{\FS{t}}\|_2^2}{2}
        \notag\\
        &
        = f(\vw_t) - r_T
        + \ip{[\nf(\vw_t)]_{\FS{t}}}{-\eta_t [\vg_t]_{\FS{t}}}
        + \frac {\eta_t^2 \beta \|[\vg_t]_{\FS{t}}\|_2^2}{2}
        \notag\\
        &
        = f(\vw_t) - r_T
        - \eta_t\ip{[\nf(\vw_t)]_{\FS{t}}}{[\nf(\vw_t)]_{\FS{t}}}
        + \eta_t \VE{2}{t}
        + \frac {\eta_t^2 \beta \|[\vg_t]_{\FS{t}}\|_2^2}{2}
        \notag\\
        &
        \leq f(\vw_t) - r_T
        + 2\lambda\eta_t(f(\vu_{t}) -r_T + r_T - f(\vw_t))
        + \eta_t \VE{2}{t}
        + \frac {\eta_t^2 \beta G^2}{2}
        \notag\\
        &
        = \left(1- 2\lambda \eta_t\right)\left(f(\vw_t) - r_T\right)
        + \frac {\eta_t^2 \beta G^2}{2}
        + \eta_t\left(\VE{1}{t} + \VE{2}{t}\right),
        \label{eq:fw_ish}
    \end{align}
    thus establishing \cref{eq:sc:1} since $2\lambda\eta_t = 2/(t+1)$.

    Next it will be proved by induction that, for any $t\geq 1$,
    \begin{equation}
        f(\vw_{t+1}) - r_T
        \leq \frac {c} {t+1}
        + \sum_{j=1}^{t} \eta_j(\VE{1}{j} + \VE{2}{j}) \prod_{l=j+1}^{t} \frac{l}{l+2},
        \label{eq:sc:2}
    \end{equation}
    where the convention $\prod_{l=j+1}^{t} l / (l+2) = 1$ is adopted for $t < j+1$.
    For the base case $t=1$, \cref{eq:sc:1} grants
    \begin{align*}
        f(\vw_{t+1}) - r_T
        &\leq
        \underbrace{\left(\frac {t-1}{t+1}\right)}_{=0}\left(f(\vw_t) - r_T\right)
        + \frac {\eta_t^2 \beta G^2}{2}
        + \eta_t\left(\VE{1}{t} + \VE{2}{t}\right)
        \\
        &\leq
        \frac {c}{t+1}
        + \sum_{j=1}^{t} \eta_j(\VE{1}{j} + \VE{2}{j}) \underbrace{\prod_{l=j+1}^{t} \frac l {l+2}}_{=1}.
    \end{align*}
    On other other hand, in the case $t > 1$, once again starting from \cref{eq:sc:1},
    \begin{align*}
        f(\vw_{t+1})
        -r_T
        &
        \leq \left(\frac{t-1}{t+1}\right)\left(f(\vw_t) - r_T\right)
        + \frac {\eta_t^2 \beta G^2}{2}
        + \eta_t\left(\VE{1}{t} + \VE{2}{t}\right)
        \\
        &
        \leq \left(\frac{t-1}{t+1}\right)\left(
            \frac {c}{t}
            + \sum_{j=1}^{t-1} \eta_j(\VE{1}{j} + \VE{2}{j}) \prod_{l=j+1}^{t-1} \frac l {l+2}
        \right)
        + \frac {\eta_t^2 \beta G^2}{2}
        + \eta_t\left(\VE{1}{t} + \VE{2}{t}\right)
        \\
        &
        = \frac {t-1} {t+1}
        \left( \frac {c}{t} + \frac {\beta G^2}{2\lambda^2(t-1)(t+1)}\right)
        + \sum_{j=1}^{t} \eta_j(\VE{1}{j} + \VE{2}{j}) \prod_{l=j+1}^{t} \frac{l}{l+2},
    \end{align*}
    thus completing the proof of \cref{eq:sc:2}.

    To simplify the error term of \cref{eq:sc:2}, note
    \begin{align*}
        j = t
        &\qquad\Longrightarrow\qquad
        \prod_{l=j+1}^{t} \frac {l}{l+2}
        = 1
        = \frac {(j+1)(j+2)}{(t+1)(t+2)},
        \\
        j = t - 1
        &\qquad\Longrightarrow\qquad
        \prod_{l=j+1}^{t} \frac {l}{l+2}
        = \frac {t}{t+2}
        = \frac {(j+1)(j+2)}{(t+1)(t+2)},
        \\
        j < t - 1
        &\qquad\Longrightarrow\qquad
        \prod_{l=j+1}^{t} \frac {l}{l+2}
        = \frac {(j+1)(j+2)}{(t+1)(t+2)}\prod_{l=j+3}^{t} \frac {l}{l}
        = \frac {(j+1)(j+2)}{(t+1)(t+2)};
    \end{align*}
    thus, for any $t\geq 1$ and $1 \leq j \leq t$,
    \[
        \eta_j \prod_{l=j+1}^{t} \frac {l}{l+2}
        = \frac {j+2}{\lambda (t+1) (t+2)}.
    \]
    Plugging this simplification back into \cref{eq:sc:2}, for $t\geq 1$,
    \begin{align}
        f(\vw_{t+1}) - r_T
        &\leq
        \frac {c} {t+1}
        + \sum_{j=1}^{t} \eta_j(\VE{1}{j} + \VE{2}{j}) \prod_{l=j+1}^{t} \frac{l}{l+2}
        \notag\\
        &=
        \frac {c} {t+1}
        + \frac {1}{\lambda (t+1)(t+2)}\sum_{j=1}^{t} (j+2)(\VE{1}{j} + \VE{2}{j}).
        \label{eq:sc:3}
    \end{align}

    Next, to instantiate the comparator $r$, consider the choice
    \[
        r_T := \frac{ \sum_{j=1}^{T} (j+2)f(\vu_{j})}{\sum_{j=1}^T (j+2)}.
    \]
    By construction, this provides
    \[
        \frac 1 {2\lambda}\sum_{j=1}^{T} (j+2) \VE{1}{j}
        =
        \sum_{j=1}^{T} (j+2) f(\vu_{j})
        - r_T \sum_{j=1}^{T} (j+2)
        =
        \sum_{j=1}^{T} (j+2) f(\vu_{j})
        - \sum_{j=1}^{T} (j+2) f(\vu_{j})
        = 0.
    \]
    Consequently, \cref{eq:sc:3} simplifies to
    \begin{equation}
        f(\vw_{T+1}) -
        \frac{ \sum_{j=1}^{T} (j+2)f(\vu_{j})}{\sum_{j=1}^T (j+2)}
        \leq
        \frac {1}{T+1}
        \left(c
            + \frac 1 {\lambda(T+2)}\sum_{j=1}^{T} (j+2)\VE{2}{j}
        \right),
        \label{eq:sc:4}
    \end{equation}
    which is the first part of the desired statement.

    For the final desired statement, it remains to control $\VE{2}{j}$ within
    \cref{eq:sc:4}.  For the expected value, let $\cF_j$ be the $\sigma$-algebra
    of information up to time $j$; then
    \begin{align*}
        \bbE\left(
            \sum_{j=1}^{T} \frac {j+2}{T+2}\VE{2}{j}
        \right)
        &=
        \bbE\left(
            \cdots
            \bbE\left(
                \bbE\left(
                    \sum_{j=1}^{T} \frac {j+2}{T+2}\VE{2}{j}
                    \Big|\cF_1
                \right)
                \Big|\cF_2
            \right)
            \cdots
        \Big| \cF_T\right)
        \\
        &=
        \sum_{j=1}^T\bbE\left(
            \frac {j+2}{T+2}\VE{2}{j}
            \Big|\cF_j
        \right)
        \\
        &= 0.
    \end{align*}
    Here the last equality holds since 
    \begin{align*} 
        \bbE\left(
            \VE{2}{j} \Big|\cF_j
        \right)
        &= \bbE\left(
            \ip{[\nf(\vw_j)]_{\FS{j}} - [\vg_j]_{\FS{j}}}{\nf(\vw_j)}
        \Big| \cF_j \right)  \\
        &= 
        \ip{
            \bbE\left(
                [\nf(\vw_j)]_{\FS{j}} - [\vg_j]_{\FS{j}}
            \Big| \cF_j \right) 
        }{\nf(\vw_j)}
         \\
         &= 0,
  \end{align*}
  which used the fact that $\nf(\vw_j)$ is constant in the
  $\sigma$-field $\cF_j$. This yields the expectation bound.        
       
    For the high probability bound, by Azuma-Hoeffding, with
    probability at least $1-\delta$, 
    \begin{align*}
        \sum_{j=1}^{T} \frac {j+2}{T+2}\VE{2}{j}
        &= \sum_{j=1}^{T}\frac {j+2}{T+2}\ip{[\nf(\vw_j)]_{\FS{j}} -
      [\vg_j]_{\FS{j}}}{\nf(\vw_j)}
        \\
        &\leq \sqrt{2 \ln(1/\delta) \sum_{j=1}^{T} 4G^4 (j+2)^2 / (T+2)^2}
        \\
        &\leq 4G^2 \sqrt{T\ln(1/\delta)}.
        \qedhere
    \end{align*}
\end{proof}

With the proof of \Cref{fact:tracking:fancy} out of the way, the proof of
\Cref{fact:tracking:fancier} follows easily.

\begin{proof}[Proof of \Cref{fact:tracking:fancier}]
    By the assumptions, $\ell(\ip{\vw}{\vx y})$ is bounded for any
    $\|\vw\|\leq D$ and $\|\vx y\|\leq X$, and since
    $\vw\mapsto \ell(\ip{\vw}{\vx y})$ is continuous, it follows that
    $\nf(\vw) = \bbE[\nabla \ell(\ip{\vw}{\vx y})]$, meaning in
    particular that
    \begin{align*}
        \|\nf(\vw)\|
        &\leq \|\bbE[\vx y\ell'(\ip{\vw}{\vx y})]\| + \lambda \|\vw\|
        \leq X + \lambda D,
        \\
        \|\nabla^2 f(\vw)\|
        &\leq \|\bbE[\vx y\ell''(\ip{\vw}{\vx y})\vx^\top y]\| + \lambda
        \leq X^2 + \lambda.
    \end{align*}
    Expanding these Lipschitz and smoothness terms
    in \Cref{fact:tracking:fancy} gives the desired result.
%
\end{proof}

\section{Summary of datasets}

Below, $n$ is the number of examples, $d$ is the number of base
features, and $s$ is the average number of non-zero base features per
example.

\begin{center}
\begin{tabular}{c||c|c|c|c}
Dataset & $n$ & $s$ & $d$ & problem
\\
\hline
\hline
20news &  18845  &  93.8854  & 101631  & binary
\\
a9a &  48841  &  13.8676  & 123  & binary
\\
abalone &  4176  &  8  & 8  & binary
\\
abalone &  4177  &  7.99952  & 10  & regression
\\
activity &  165632  &  18.5489  & 20  & binary
\\
adult &  48842  &  11.9967  & 105  & binary
\\
bio &  145750  &  73.4184  & 74  & binary
\\
cal-housing &  20639  &  8  & 8  & regression
\\
census &  299284  &  32.0072  & 401  & binary
\\
comp-activ-harder &  8191  &  11.5848  & 12  & regression
\\
covtype &  581011  &  11.8789  & 54  & binary
\\
cup98-target &  95411  &  310.982  & 10825  & binary
\\
eeg-eye-state &  14980  &  13.9901  & 14  & binary
\\
ijcnn1&  24995  &  13  & 22  & binary
\\
kdda&  8407751  &  36.349  & 19306083  & binary
\\
kddcup2009 &  50000  &  58.4353  & 71652  & binary
\\
letter &  20000  &  15.5807  & 16  & binary
\\
magic04 &  19020  &  9.98728  & 10  & binary
\\
maptaskcoref &  158546  &  40.4558  & 5944  & binary
\\
mushroom &  8124  &  22  & 117  & binary
\\
nomao &  34465  &  82.3306  & 174  & binary
\\
poker&  946799  &  10  & 10  & binary
\\
rcv1&  781265  &  75.7171  & 43001  & binary
\\
shuttle&  43500  &  7.04984  & 9  & binary
\\
skin &  245057  &  2.948  & 3  & binary
\\
slice&  53500  &  134.575  & 384  & regression
\\
titanic &  2201  &  3  & 8  & binary
\\
vehv2binary &  299254  &  48.5652  & 105  & binary
\\
w8a &  49749  &  11.6502  & 300  & binary
\\
year&  463715  &  90  & 90  & regression
\end{tabular}
\end{center}

\section{Further experimental results}

We will show three more sets of results in the appendix. The first set
contains a bar plot detailing the performance of the \lalg, \qalg and
\calg baselines, as well as \alg with $\alpha \in \{0.125, 0.25, 0.5,
0.75, 1\}$ on all of our 30 datasets. For each method, we present the
relative error~\eqref{eq:rel-error} in Figure~\ref{fig:full-error}.

Since the statistical error by itself only tells half the story, we
also include a similar plot for relative running times in
Figure~\ref{fig:full-time}. 

Finally, we also want to highlight that despite the competitive
statistical performance, our adaptive methods indeed generate a much
smaller number of monomials. To this end, we compute the average
number of non-zero features per example on all datasets for all
methods. These plots are presented in Figure~\ref{fig:nfeatures}, with
the same color coding used for algorithms as the previous two plots. 

We include tables with all the (non-relative) errors and times in
Table~\ref{table:summary1} and
Table~\ref{table:summary2}.

\begin{landscape}

\begin{figure}
\centering
\includegraphics[height=\textheight]{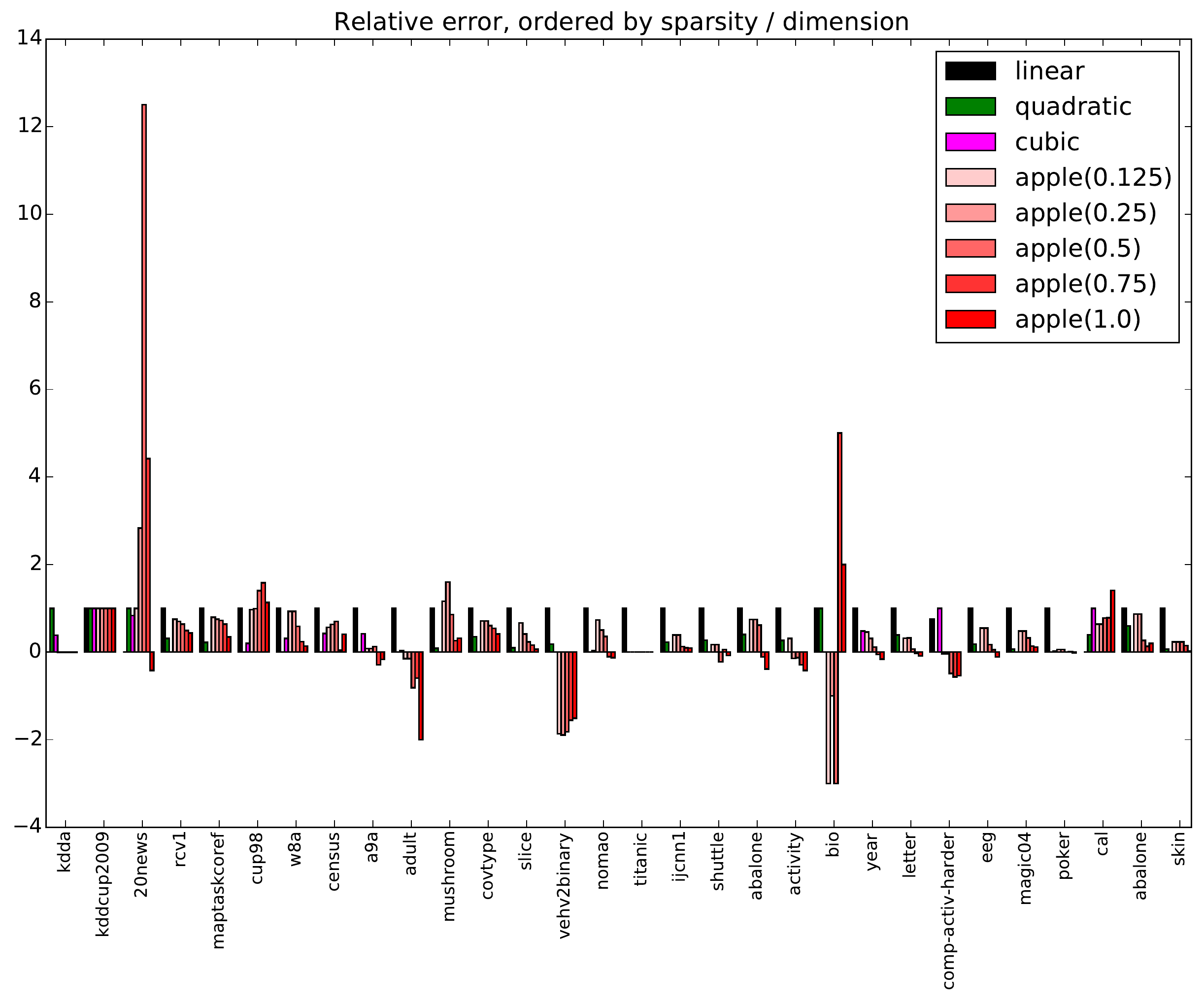}
\caption{Relative error plots for \alg and baselines on all 30
  datasets. Should be viewed in color.}
\label{fig:full-error}
\end{figure}

\begin{figure}
\centering
\includegraphics[height=\textheight]{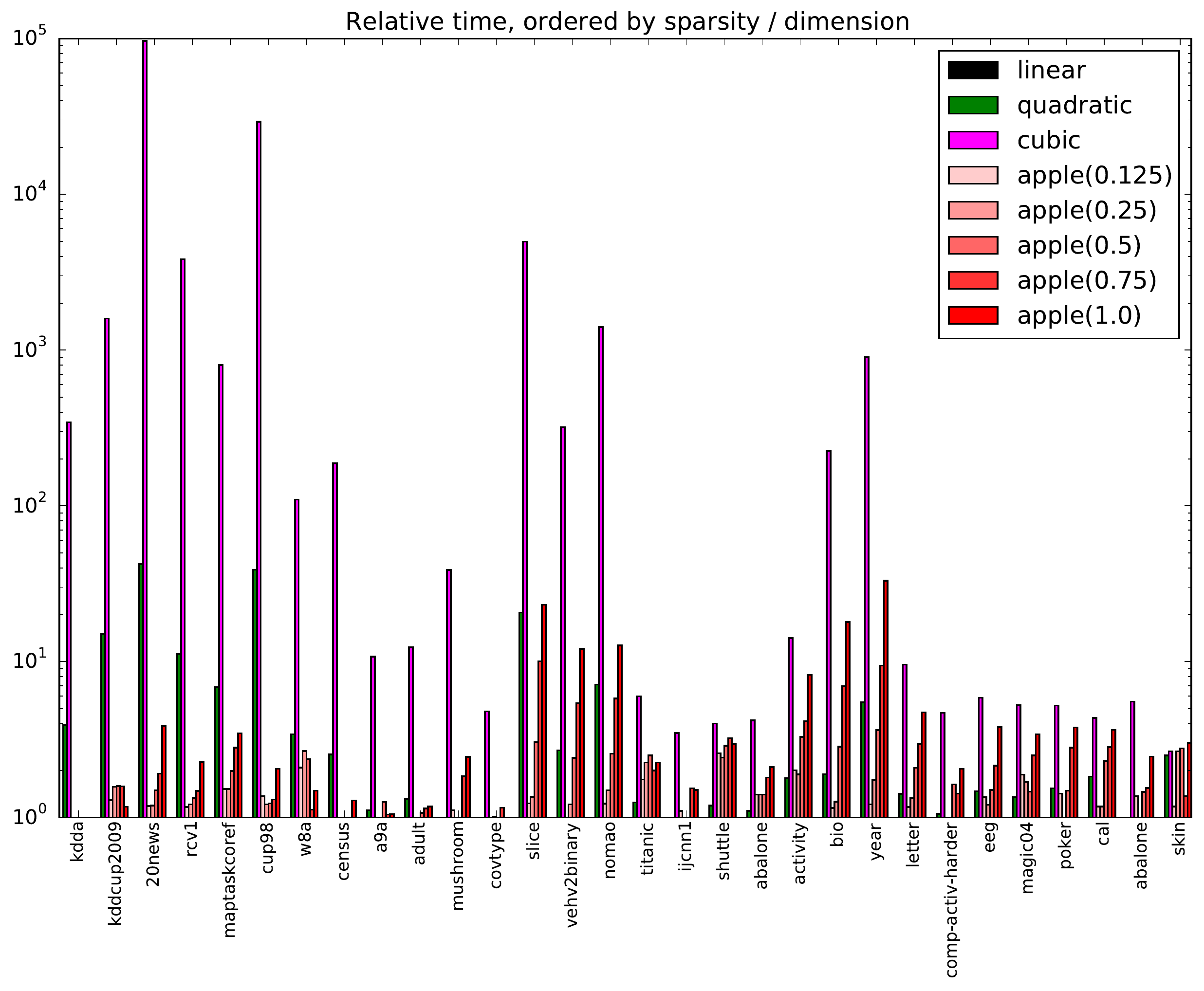}
\caption{Relative time plots for \alg and baselines on all 30
  datasets. Should be viewed in color.}
\label{fig:full-time}
\end{figure}

\begin{figure}
\centering
\includegraphics[height=\textheight]{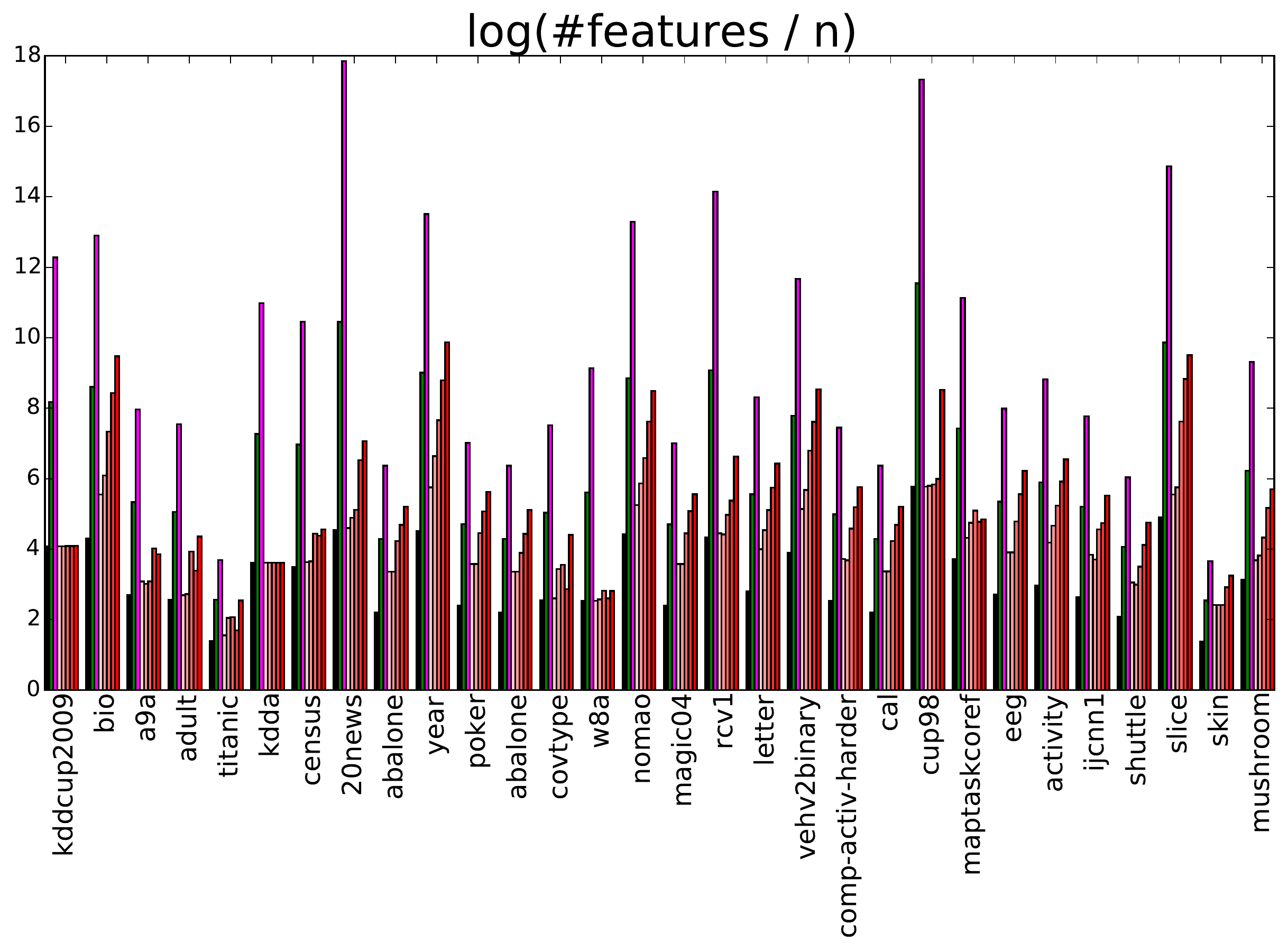}
\caption{Relative time plots for \alg and baselines on all 30
  datasets. Should be viewed in color.}
\label{fig:nfeatures}
\end{figure}

\begin{table}
  \centering
  \begin{small}
  \begin{tabular} {|c|c|c|c|c|c|c|c|c|} \hline 
~& \lalg & \qalg & \calg & \alg (1) & \alg (0.75) & \alg (0.5) & \alg (0.25) & \alg (0.125) \\ \hline 
 \multirow{2}{*}{bio} & 3.122e-03 & 3.122e-03 & 3.087e-03 & 2.985e-03 & 3.053e-03 & 2.985e-03 & 3.259e-03 & 3.156e-03  \\ 
& 2.644e+00  & 5.000e+00  & 5.965e+02  & 3.036e+00  & 3.340e+00  & 7.516e+00  & 1.841e+01  & 4.745e+01   \\ \hline
 \multirow{2}{*}{a9a} & 1.510e-01 & 1.485e-01 & 1.496e-01 & 1.488e-01 & 1.488e-01 & 1.489e-01 & 1.478e-01 & 1.481e-01  \\ 
& 3.880e-01  & 4.320e-01  & 4.176e+00  & 2.840e-01  & 3.000e-01  & 4.880e-01  & 4.040e-01  & 4.080e-01   \\ \hline
 \multirow{2}{*}{adult} & 1.557e-01 & 1.529e-01 & 1.531e-01 & 1.525e-01 & 1.525e-01 & 1.507e-01 & 1.513e-01 & 1.474e-01  \\ 
& 3.440e-01  & 4.520e-01  & 4.252e+00  & 2.480e-01  & 2.400e-01  & 3.680e-01  & 3.920e-01  & 4.040e-01   \\ \hline
 \multirow{2}{*}{titanic} & 2.182e-01 & 2.136e-01 & 2.136e-01 & 2.136e-01 & 2.136e-01 & 2.136e-01 & 2.136e-01 & 2.136e-01  \\ 
& 1.600e-02  & 2.000e-02  & 9.601e-02  & 2.800e-02  & 3.600e-02  & 4.000e-02  & 3.200e-02  & 3.600e-02   \\ \hline
 \multirow{2}{*}{kdda} & 1.240e-01 & 1.272e-01 & 1.253e-01 & 1.240e-01 & 1.240e-01 & 1.240e-01 & 1.240e-01 & 1.240e-01  \\ 
& 9.492e+01  & 3.715e+02  & 3.266e+04  & 7.629e+01  & 6.689e+01  & 8.318e+01  & 5.768e+01  & 8.463e+01   \\ \hline
 \multirow{2}{*}{census} & 4.748e-02 & 4.579e-02 & 4.651e-02 & 4.674e-02 & 4.686e-02 & 4.698e-02 & 4.586e-02 & 4.648e-02  \\ 
& 3.068e+00  & 7.784e+00  & 5.754e+02  & 2.200e+00  & 2.180e+00  & 2.144e+00  & 2.532e+00  & 3.936e+00   \\ \hline
 \multirow{2}{*}{20news} & 8.119e-02 & 8.437e-02 & 8.384e-02 & 8.437e-02 & 9.021e-02 & 1.210e-01 & 9.525e-02 & 7.986e-02  \\ 
& 5.440e-01  & 2.303e+01  & 5.262e+04  & 6.440e-01  & 6.480e-01  & 8.121e-01  & 1.040e+00  & 2.112e+00   \\ \hline
 \multirow{2}{*}{abalone\_bin} & 2.898e-01 & 2.826e-01 & 2.719e-01 & 2.874e-01 & 2.874e-01 & 2.766e-01 & 2.743e-01 & 2.754e-01  \\ 
& 4.400e-02  & 4.000e-02  & 2.440e-01  & 6.000e-02  & 4.400e-02  & 6.400e-02  & 6.800e-02  & 1.080e-01   \\ \hline
 \multirow{2}{*}{year} & 1.157e-02 & 1.073e-02 & 1.113e-02 & 1.112e-02 & 1.099e-02 & 1.083e-02 & 1.069e-02 & 1.060e-02  \\ 
& 1.261e+01  & 6.915e+01  & 1.136e+04  & 1.529e+01  & 2.201e+01  & 4.585e+01  & 1.189e+02  & 4.179e+02   \\ \hline
 \multirow{2}{*}{poker} & 4.555e-01 & 4.091e-01 & 4.100e-01 & 4.119e-01 & 4.119e-01 & 4.092e-01 & 4.099e-01 & 4.085e-01  \\ 
& 4.388e+00  & 6.736e+00  & 2.294e+01  & 6.228e+00  & 3.836e+00  & 6.516e+00  & 1.230e+01  & 1.657e+01   \\ \hline
 \multirow{2}{*}{abalone\_reg} & 8.052e+00 & 7.489e+00 & 7.107e+00 & 7.812e+00 & 7.812e+00 & 7.690e+00 & 7.003e+00 & 6.740e+00  \\ 
& 4.000e-02  & 4.400e-02  & 1.680e-01  & 5.600e-02  & 5.600e-02  & 5.600e-02  & 7.200e-02  & 8.400e-02   \\ \hline
 \multirow{2}{*}{kddcup2009} & 7.310e-02 & 7.310e-02 & 7.310e-02 & 7.310e-02 & 7.310e-02 & 7.310e-02 & 7.310e-02 & 7.310e-02  \\ 
& 7.600e-01  & 1.144e+01  & 1.213e+03  & 9.801e-01  & 1.196e+00  & 1.208e+00  & 1.204e+00  & 8.881e-01   \\ \hline
 \multirow{2}{*}{covtype} & 2.450e-01 & 2.184e-01 & 2.039e-01 & 2.331e-01 & 2.331e-01 & 2.287e-01 & 2.261e-01 & 2.208e-01  \\ 
& 4.444e+00  & 3.116e+00  & 2.133e+01  & 4.204e+00  & 4.488e+00  & 2.856e+00  & 5.120e+00  & 4.192e+00   \\ \hline
 \multirow{2}{*}{w8a} & 1.538e-02 & 1.246e-02 & 1.337e-02 & 1.518e-02 & 1.518e-02 & 1.417e-02 & 1.317e-02 & 1.286e-02  \\ 
& 1.320e-01  & 4.520e-01  & 1.448e+01  & 2.760e-01  & 3.520e-01  & 3.120e-01  & 1.480e-01  & 1.960e-01   \\ \hline
 \multirow{2}{*}{nomao} & 6.325e-02 & 5.063e-02 & 5.107e-02 & 5.992e-02 & 5.701e-02 & 5.513e-02 & 4.933e-02 & 4.904e-02  \\ 
& 5.000e-01  & 3.564e+00  & 7.038e+02  & 6.120e-01  & 7.480e-01  & 1.280e+00  & 2.900e+00  & 6.348e+00   \\ \hline
\end{tabular}
  \end{small}
  \caption{Errors (first row) and running time in seconds (second row)
  for each dataset for the baselines and \alg variants (first 15 datasets).}
  \label{table:summary1}
\end{table}

\begin{table}
  \centering
  \begin{small}
  \begin{tabular} {|c|c|c|c|c|c|c|c|c|} \hline 
~& \lalg & \qalg & \calg & \alg (1) & \alg (0.75) & \alg (0.5) & \alg (0.25) & \alg (0.125) \\ \hline 
 \multirow{2}{*}{magic04} & 2.142e-01 & 1.672e-01 & 1.638e-01 & 1.880e-01 & 1.880e-01 & 1.801e-01 & 1.706e-01 & 1.696e-01  \\ 
& 1.040e-01  & 1.400e-01  & 5.480e-01  & 1.960e-01  & 1.760e-01  & 1.520e-01  & 2.600e-01  & 3.560e-01   \\ \hline
 \multirow{2}{*}{rcv1} & 4.860e-02 & 4.060e-02 & 3.701e-02 & 4.570e-02 & 4.511e-02 & 4.438e-02 & 4.273e-02 & 4.205e-02  \\ 
& 1.627e+01  & 1.823e+02  & 6.251e+04  & 1.895e+01  & 1.977e+01  & 2.171e+01  & 2.401e+01  & 3.671e+01   \\ \hline
 \multirow{2}{*}{letter} & 2.273e-01 & 1.918e-01 & 1.688e-01 & 1.872e-01 & 1.878e-01 & 1.727e-01 & 1.670e-01 & 1.638e-01  \\ 
& 1.440e-01  & 2.040e-01  & 1.376e+00  & 1.680e-01  & 1.920e-01  & 3.000e-01  & 4.280e-01  & 6.800e-01   \\ \hline
 \multirow{2}{*}{vehv2binary} & 3.400e-02 & 2.670e-02 & 2.505e-02 & 8.337e-03 & 8.087e-03 & 8.772e-03 & 1.109e-02 & 1.151e-02  \\ 
& 3.364e+00  & 9.065e+00  & 1.079e+03  & 2.880e+00  & 4.076e+00  & 8.105e+00  & 1.825e+01  & 4.065e+01   \\ \hline
 \multirow{2}{*}{comp} & 3.409e-03 & 2.627e-03 & 3.662e-03 & 2.587e-03 & 2.587e-03 & 2.124e-03 & 2.038e-03 & 2.070e-03  \\ 
& 7.601e-02  & 8.000e-02  & 3.560e-01  & 6.401e-02  & 6.400e-02  & 1.240e-01  & 1.080e-01  & 1.560e-01   \\ \hline
 \multirow{2}{*}{cal\_housing} & 7.410e-02 & 8.651e-02 & 1.055e-01 & 9.414e-02 & 9.414e-02 & 9.856e-02 & 9.881e-02 & 1.183e-01  \\ 
& 9.201e-02  & 1.680e-01  & 4.000e-01  & 1.080e-01  & 1.080e-01  & 2.120e-01  & 2.600e-01  & 3.360e-01   \\ \hline
 \multirow{2}{*}{cup98} & 5.697e-02 & 3.841e-02 & 4.215e-02 & 5.646e-02 & 5.687e-02 & 6.450e-02 & 6.780e-02 & 5.948e-02  \\ 
& 3.452e+00  & 1.342e+02  & 1.010e+05  & 4.736e+00  & 4.180e+00  & 4.240e+00  & 4.504e+00  & 7.084e+00   \\ \hline
 \multirow{2}{*}{maptaskcoref} & 1.087e-01 & 7.553e-02 & 6.598e-02 & 9.997e-02 & 9.805e-02 & 9.691e-02 & 9.338e-02 & 8.083e-02  \\ 
& 9.361e-01  & 6.412e+00  & 7.504e+02  & 1.424e+00  & 1.424e+00  & 1.856e+00  & 2.628e+00  & 3.240e+00   \\ \hline
 \multirow{2}{*}{eeg\_eye\_state} & 3.815e-01 & 2.573e-01 & 2.300e-01 & 3.127e-01 & 3.127e-01 & 2.557e-01 & 2.383e-01 & 2.136e-01  \\ 
& 1.360e-01  & 2.000e-01  & 7.960e-01  & 1.840e-01  & 1.640e-01  & 2.040e-01  & 2.920e-01  & 5.160e-01   \\ \hline
 \multirow{2}{*}{activity} & 1.298e-02 & 8.422e-03 & 6.762e-03 & 8.694e-03 & 5.856e-03 & 5.977e-03 & 4.951e-03 & 4.166e-03  \\ 
& 5.560e-01  & 9.921e-01  & 7.864e+00  & 1.116e+00  & 1.048e+00  & 1.828e+00  & 2.304e+00  & 4.564e+00   \\ \hline
 \multirow{2}{*}{ijcnn1} & 7.942e-02 & 4.501e-02 & 3.481e-02 & 5.221e-02 & 5.221e-02 & 4.041e-02 & 3.921e-02 & 3.901e-02  \\ 
& 2.000e-01  & 1.800e-01  & 6.960e-01  & 2.200e-01  & 1.800e-01  & 1.840e-01  & 3.080e-01  & 3.000e-01   \\ \hline
 \multirow{2}{*}{shuttle} & 2.644e-02 & 1.391e-02 & 9.195e-03 & 1.218e-02 & 1.218e-02 & 5.402e-03 & 1.023e-02 & 7.931e-03  \\ 
& 1.040e-01  & 1.240e-01  & 4.160e-01  & 2.680e-01  & 2.520e-01  & 3.000e-01  & 3.360e-01  & 3.080e-01   \\ \hline
 \multirow{2}{*}{slice} & 7.243e-03 & 1.403e-03 & 7.797e-04 & 5.095e-03 & 3.455e-03 & 2.302e-03 & 1.825e-03 & 1.196e-03  \\ 
& 1.600e+00  & 3.303e+01  & 7.944e+03  & 1.972e+00  & 2.168e+00  & 4.880e+00  & 1.609e+01  & 3.709e+01   \\ \hline
 \multirow{2}{*}{skin} & 7.421e-02 & 1.045e-02 & 5.835e-03 & 2.165e-02 & 2.165e-02 & 2.165e-02 & 1.585e-02 & 7.366e-03  \\ 
& 4.600e-01  & 1.148e+00  & 1.220e+00  & 5.400e-01  & 1.220e+00  & 1.276e+00  & 6.280e-01  & 1.384e+00   \\ \hline
 \multirow{2}{*}{mushroom} & 5.723e-02 & 5.538e-03 & 6.154e-04 & 6.646e-02 & 9.108e-02 & 4.923e-02 & 1.538e-02 & 1.846e-02  \\ 
& 7.200e-02  & 6.800e-02  & 2.788e+00  & 8.001e-02  & 5.600e-02  & 6.000e-02  & 1.320e-01  & 1.760e-01   \\ \hline
\end{tabular}
  \end{small}
  \caption{Errors (first row) and running time in seconds (second row)
  for each dataset for the baselines and \alg variants (last 15 datasets).}
  \label{table:summary2}
\end{table}

\end{landscape}

\end{document}